%% file: example.tex
\documentclass{article}

\usepackage[dvipsnames]{xcolor}

\usepackage[final]{corl_2026} %
\hypersetup{
  pdftitle={Walk Like This: Stylized Locomotion for Physics-Based Characters through Example-Driven Diffusion},
  pdfauthor={Joshua Aurand, Agon Serifi, Sammy Christen, Ruben Grandia, Espen Knoop, Marco Hutter, Moritz Bächer}
}

\title{Walk Like This: Stylized Locomotion for Physics- Based Characters through Example-Driven Diffusion}

\usepackage{booktabs} %
\usepackage[normalem]{ulem}
\usepackage{siunitx}
\usepackage{multirow}
\usepackage{multicol}
\usepackage{subcaption}
\usepackage{nicefrac}
\usepackage{bbm}
\usepackage{bm}
\usepackage{diagbox}
\usepackage{wrapfig}
\usepackage{amssymb}
\usepackage{amsmath}
\usepackage{mathtools}
\usepackage{pifont}   %
\usepackage{tabularx} %

\usepackage{caption}
\usepackage{subcaption}
\usepackage{graphicx}
\usepackage[export]{adjustbox} %
\usepackage{xcolor}

\usepackage[normalem]{ulem}

\newcommand\Fig[1]{Fig.~\ref{fig:#1}}
\newcommand\Sec[1]{Sec.~\ref{sec:#1}}
\newcommand\Eq[1]{Eq.~\eqref{eq:#1}}
\newcommand\Tab[1]{Tab.~\ref{tab:#1}}
\newsavebox{\mytablebox}

\author{
  \textbf{Joshua Aurand}\textsuperscript{1,2} \quad
  \textbf{Agon Serifi}\textsuperscript{1} \quad
  \textbf{Sammy Christen}\textsuperscript{1} \quad
  \textbf{Ruben Grandia}\textsuperscript{1}\\[2pt]
  \textbf{Espen Knoop}\textsuperscript{1} \quad
  \textbf{Marco Hutter}\textsuperscript{2} \quad
  \textbf{Moritz B\"acher}\textsuperscript{1}\\[4pt]
  \textsuperscript{1}Disney Research Z\"urich, Switzerland \qquad
  \textsuperscript{2}ETH Z\"urich, Switzerland\\[2pt]
  \texttt{joshua.aurand@disneyresearch.com}
}

\begin{document}
\maketitle

\vspace{-5mm}
\begin{figure}[h!]
    \centering
    \includegraphics[width=\linewidth]{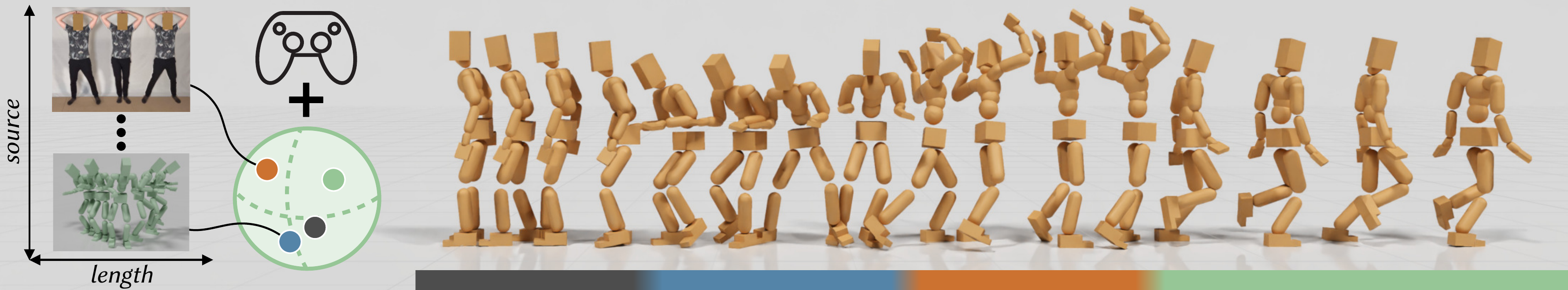} \\
    \caption{We introduce a real-time controller for physics-based characters that lets users steer the character with joystick commands while conditioning on style from unseen examples of arbitrary duration and various sources (e.g., MoCap, video).} 
  \label{fig:teaser}
\end{figure}
\begin{abstract}
We present a real-time framework for expressive legged locomotion that enables independent control over velocity and style. 
Unlike existing data-driven controllers, which typically require explicit style labels for training, our method efficiently leverages large-scale unstructured motion data and generalizes to unseen styles. 
We disentangle style from functional locomotion through a tailored sampling strategy used to train a style encoder whose output conditions an autoregressive diffusion model. During inference, the user provides a style reference clip and a joystick velocity command. The diffusion model then generates a kinematic reference tracked by a reinforcement learning controller. 
To improve user controllability and stylization strength, we propose an attention biasing mechanism and a velocity command correction strategy. 
We demonstrate our method on a physics-based humanoid character in simulation and on a physical robot, showcasing robust real-time control and continuous style modulation. 
The results can be best seen in our \href{https://www.youtube.com/watch?v=roFnS6N7nJ4}{supplemental video}\footnote{\url{https://www.youtube.com/watch?v=roFnS6N7nJ4}}.

\end{abstract}


\section{Introduction}

A central challenge in physics-based humanoid control is to move beyond functional locomotion and to perform tasks in an expressive gait \emph{style}. Although some stylized locomotion controllers exist, they are typically designed around fixed objectives or limited to imitating a narrow distribution of motions \citep{mu2026smp, peng_amp_2021, peng_ase_2022, grandia_design_2024, piseno_stylized_2025}, offering little flexibility for stylistic variation and generalization after training. Similarly, while CAMDM~\citep{chen_taming_2024} achieves real-time stylized character control, its conditioning on explicit style labels prevents generalization to new styles. Style transfer offers an alternative by mapping style onto target motions \citep{aberman_2020_unpaired, jang_2022_motionpuzzle, zhong_2024_smoodi}, but these methods are not designed for real-time robotic tracking or interactive control. On the other hand, motion generators~\citep{serifi_robot_2024,tevet_closd_2024, yuan_physdiff_2023, xu_parc_2025} can be grounded in physics but lack expressive control over style. Hence, a real-time framework that adapts to unseen styles on humanoid robots is currently missing.

To fill this gap, we propose a framework for the interactive control of humanoid robots that enables real-time user control via joystick commands and an example motion clip that contains the desired style characteristics. This gives users intuitive control over both \emph{what} the robot does and \emph{how} it performs these movements. To enable this, we build a stylized generative motion model based on autoregressive diffusion \cite{chen_taming_2024}. Crucially, we condition the model on a style embedding that is extracted from a style example through a transformer-based encoder. We jointly train the diffusion model and style encoder on unlabeled data, and use a tailored sampling strategy to disentangle style from functional locomotion. Our method both generalizes to new styles and gives users the flexibility to provide style examples of any duration. We highlight both aspects in our experiments by extracting a style directly from a recorded human video demonstration.

To enhance style adherence and enable rapid style transitions, we introduce an inference-time technique that re-weights the attention scores within the diffusion model, outperforming standard classifier-free guidance (CFG) \citep{ho2022classifierfreediffusionguidance} both in efficiency and effectiveness. Furthermore, we add a velocity correction scheme that mitigates drift when following joystick commands. Finally, we bridge the gap between generated motion and hardware dynamics by pairing the diffusion model with a reinforcement learning (RL) tracking controller. This combines the \emph{expressive power} of a motion generator with the \emph{robustness} of an RL low-level controller, as we validate with closed-loop simulation and hardware deployment. 

Our core contributions: \textbf{(i) A generative motion-control framework} enabling user control over locomotion and style, trained from large-scale unstructured datasets and demonstrated on hardware. \textbf{(ii) An autoregressive diffusion model and style encoder} that disentangle style from functional locomotion through an effective sampling strategy during training. \textbf{(iii) An inference-time attention biasing technique} to modulate stylization strength alongside \textbf{a velocity correction scheme} to mitigate drift. We validate each design choice through ablations.

\section{Related Work}
\definecolor{Green4}{RGB}{1,120,32}
\newcommand{\cmark}{\ding{51}} %
\newcommand{\xmark}{\ding{55}} %
\newcommand{\greencmark}{\color{Green4}{\cmark}}
\newcommand{\redxmark}{\color{red}{\xmark}}
\newcommand{\mmark}{\textcolor{orange!90!black}{\ensuremath{\thicksim}}}

\begin{table}
    \vspace{-15pt} %
    \centering
    \small
    \captionsetup{type=table} %
    \caption{\textbf{Comparison to Related Work.} Real-time (RT), generative (Gen), unseen styles (Uns Sty), physics-based model (Phys), joystick control (Joy Ctrl). {\textit{Legend: }\greencmark~Yes~~~\mmark~\textcolor{orange!90!black}{Partially}~~~\redxmark~No}}
    \label{tab:comparison}
    \setlength{\tabcolsep}{4pt} %
     \begin{tabular}{l c c c c c}
        \toprule
        \textbf{Approach} & \textbf{RT} & \textbf{Gen} & \textbf{Uns Sty}  & \textbf{Phys} & \textbf{Joy Ctrl} \\
        \midrule
        Style Transfer \citep{aberman_2020_unpaired, jang_2022_motionpuzzle, zhong_2024_smoodi, song_arbitrary_motion_style_transfer, kim2024most, 10.1145/3757376.3771387, raab_monkey_2024, zhong_2025_smoogpt}   & \redxmark   & \mmark & \greencmark & \redxmark & \redxmark \\
        Implicit Rewards \citep{mu2026smp,peng_amp_2021} & \greencmark & \mmark & \redxmark & \greencmark & \mmark  \\
        ``Physical Diffusion'' \citep{serifi_robot_2024, tevet_closd_2024, yuan_physdiff_2023, xu_parc_2025} & \mmark & \greencmark & \redxmark & \greencmark & \mmark   \\
        CAMDM \citep{chen_taming_2024}            & \greencmark & \greencmark & \redxmark  & \redxmark & \greencmark  \\
        \midrule
        \textbf{Ours}     & \greencmark & \greencmark & \greencmark & \greencmark & \greencmark \\
        \bottomrule
    \end{tabular} 
\end{table}

Our work builds upon advancements in kinematic motion synthesis and physics-based control. \Tab{comparison} contrasts our work with prior approaches.

\paragraph{Stylized Motion Generation}
Stylized motion generation traditionally relies on \textit{style transfer} from reference to content motion \citep{10.1145/2766999}, with modern approaches utilizing network activation matching \citep{aberman_2020_unpaired, song_arbitrary_motion_style_transfer, kim2024most, 10.1145/3757376.3771387} or diffusion models \citep{zhong_2024_smoodi, raab_monkey_2024, zhong_2025_smoogpt}. While they produce high-quality results, these methods are usually too slow for real-time use and depend heavily on the source content motion. Even when coupled with generative models to mitigate this dependency \citep{zhong_2024_smoodi, zhong_2025_smoogpt}, they still lack the performance and controllability required for interactive applications. In contrast, purely generative approaches are better suited for real-time interaction \citep{chen_taming_2024, 10.1145/3414685.3417836, starke_deepphase_2022, mason_2022_real, 10.1145/3197517.3201366, starke_categorical_2024, shi_interactive_2024}. Closest to our work is CAMDM \cite{chen_taming_2024}, a state-of-the-art interactive generator that uses an autoregressive diffusion model conditioned on discrete labels. While CAMDM achieves real-time, high-quality synthesis, it is restricted to seen training styles, provides purely kinematic motion, and lacks explicit control over style intensity or transitions. We overcome these limitations by introducing a more general style conditioning mechanism, explicit control over stylization strength, and an RL tracker to enforce physicality.

\paragraph{Physics-Based Character Control}
Physics-based RL is well-established for robust humanoid locomotion \citep{ radosavovic_real-world_2024, gu_advancing_2024, zhuang_humanoid_2024, li_reinforcement_2025, luo_sonic_2025}. Compared to command-based strategies \citep{radosavovic_real-world_2024, gu_advancing_2024, 11128333}, tracking-based approaches offer superior style control \citep{luo_sonic_2025, serifi_vmp_2024, peng_deepmimic_2018}, scaling from single references \citep{peng_deepmimic_2018} to large datasets via mixture-of-experts \cite{won_scalable_2020, luo_perpetual_2023} or implicit rewards \citep{mu2026smp, peng_amp_2021, peng_ase_2022}. However, frameworks like CALM \citep{tessler_calm_2023} fail to explicitly disentangle style from skills, and character-specific gait-design tools \cite{grandia_design_2024, piseno_stylized_2025, muller_olaf_2025, hopkins_interactive_2024} lack generalization. To enforce physical feasibility, recent trends combine kinematic diffusion with RL through iterative training \cite{xu_parc_2025}, joint state-action prediction \cite{huang_diffuse-cloc_2025}, or closed-loop fine-tuning \cite{tevet_closd_2024}, though these are often limited to simulation. RobotMDM \citep{serifi_robot_2024} tackles the sim-to-real gap by fine-tuning the diffusion model. We instead train a tracking policy from scratch on a large dataset of diffusion-generated motions.

\section{Background}
\label{sec:background}
We build on the framework of Motion Diffusion Models (MDMs) \citep{tevet_human_2022, zhang_motiondiffuse_2024}. Given a noisy motion $\mathbf{M}^d$, whose noise level depends on the diffusion step $d$, an MDM $\mathcal{M}$ predicts the clean motion $\mathbf{M}^0$. Particularly relevant to our work is CAMDM \citep{chen_taming_2024}, an autoregressive MDM that predicts the next $F$ frames of motion $\mathbf{M}^0_\text{fut}$, conditioned on the past $P$ frames of motion $\mathbf{M}_\text{pst}$, as well as future root trajectory $\mathbf{C}_\text{fut}$ and a discrete style label $\ell_\text{syl}$:
\begin{equation}
    \mathbf{M}^0_\text{fut} = \mathcal{M}(\mathbf{M}^d_\text{fut}, d; \mathbf{M}_\text{pst}, \mathbf{C}_\text{fut}, \ell_{\text{syl}}).
\end{equation}
The future root trajectory contains $x-y$ coordinates and the root orientation of the character in all $F$ future frames, computed by integrating world-frame velocity commands. $\mathcal{M}$ itself is a transformer-encoder \citep{vaswani_attention_2017} where all the inputs are tokenized individually. During inference, a pure noise sample is gradually denoised over multiple steps to reach the final clean prediction. We refer to \citep{chen_taming_2024, tevet_human_2022} for details.

\section{Method}

\begin{figure}
    \centering
    \includegraphics{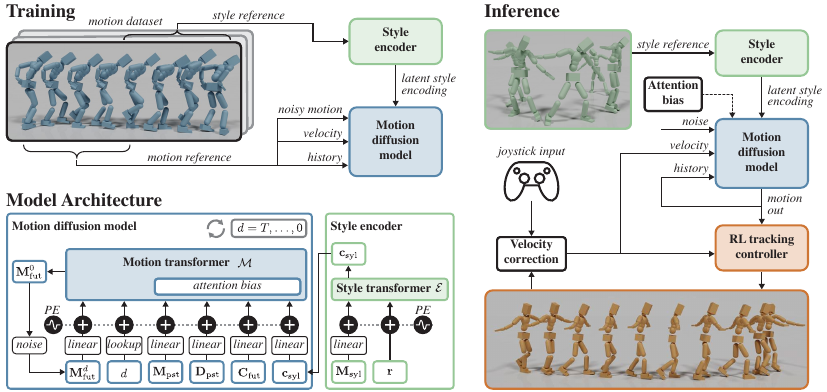}
    \caption{\textbf{Overview.} We \textit{train} an autoregressive diffusion model on unstructured motion datasets. We explicitly encode style in the style encoder, enabling generalization to unseen style references of varying length at inference. During \textit{inference}, users can control the robot's velocity through joystick commands, while selecting a desired style clip. The motion diffusion model's output is autoregressively rolled out and passed to an RL-based tracking controller.}
    \label{fig:overview}
    \vspace{-3mm}
\end{figure}

We develop a real-time interactive controller for stylized locomotion of physics-based characters, as illustrated in \Fig{overview}. Our framework contains two models: a kinematic, autoregressive diffusion model (\Sec{diffusion}) and an RL-based tracking controller (\Sec{RL}). 

The diffusion model generates kinematic motions based on user velocity commands and a stylized motion clip. A style encoder learns to extract the style from the motion in a compressed latent space. Both the generator and the style encoder are trained jointly from an unstructured motion dataset. During inference, we propose two simple yet effective techniques to improve stylization and adherence to user commands (\Sec{inference}). First, we modify bias terms in the generator transformer's attention layers. Second, we enable intuitive user control through a velocity correction module that mitigates global drift. 
To transfer the generated kinematic motions onto a physics-based character, we use an RL-based tracking controller. We train this controller on a large dataset generated through autoregressive rollouts of our diffusion model. 

\subsection{Kinematic Diffusion Model}
\label{sec:diffusion}
We use an autoregressive diffusion model for predicting the character's next $F$ states $\mathbf{M}_\text{fut}$, inspired by CAMDM \citep{chen_taming_2024} (see \Sec{background}). We condition the model on the past $P$ states $\mathbf{M}_\text{pst}$ and motion differences $\mathbf{D}_{\text{pst}}$, future root states $\mathbf{C}_\text{fut}$ and a style condition $\mathbf{c}_\text{syl}$:
\begin{equation}
\mathbf{M}^0_{\text{fut}} = \mathcal{M}(\mathbf{M}^d_{\text{fut}}, d ; \mathbf{M}_{\text{pst}}, \mathbf{D}_{\text{pst}}, \mathbf{C}_{\text{fut}}, \mathbf{c}_{\text{syl}}),
\label{eq:diffusion}
\end{equation}
where $\mathcal{M}$ is a transformer-encoder. Inputs are tokenized individually per timestep through linear maps before being fed to the transformer, except the diffusion step $d$, for which we use a learned lookup table. We use a motion representation similar to \citep{chen_taming_2024}, where the state at each timestep consists of the root position, orientation, joint orientations as well as link positions and orientations (details can be found in Appendix \ref{sec:app_diffusion}). 
Unlike CAMDM, the future root states $\mathbf{C}_{\text{fut}}$ contain desired root yaw angles and not full rotations so that the character's roll and pitch are fully controlled through the imposed style. Furthermore, we add past motion differences $\mathbf{D}_\text{pst}$ as an additional condition, computed as position and rotation differences of consecutive frames in $\mathbf{M}_{\text{pst}}$. While the information in $\mathbf{D}_\text{pst}$ is redundant, we found it to improve coherence during autoregressive rollouts, e.g., consistent swing directions of arms.

\paragraph{Encoding Style}
We compute style embeddings $\mathbf{c}_\text{syl}$ by compressing a stylized motion clip $\mathbf{M}_\text{syl}$ to a latent representation using our style transformer $\mathcal{E}$, i.e., $\mathbf{c}_\text{syl} = \mathcal{E}(\mathbf{M}_\text{syl})$. Contrary to discrete style labels~\cite{chen_taming_2024}, such inputs contain explicit information about the style and allow generalization outside the set of training styles \citep{aberman_2020_unpaired}. The style transformer $\mathcal{E}$ is also a transformer-encoder, where all frames of the style motion are tokenized through a linear projection. The transformer architecture is well suited to represent the various timespans $S$ of different styles. In addition, we enforce $\|\mathbf{c}_\text{syl} \| = 1$ through normalization which facilitates generalization \citep{peng_ase_2022, karras2019stylebasedgeneratorarchitecturegenerative} and allows mixing styles through spherical linear interpolation. Inspired by the use of register tokens in vision transformers \citep{darcet2024visiontransformersneedregisters}, we use an input-independent register token $\mathbf{r}$ in the inputs of the style transformer. It serves as a working space to capture relevant style information and we use its value after the final transformer layer as the output $\mathbf{c}_\text{syl}$. See \Fig{overview} for a visualization of the style encoder in the full model architecture.

\paragraph{Training}
Both $\mathcal{M}$ and $\mathcal{E}$ are trained jointly. We first sample a motion sequence from the training data and then sample $P+F$ consecutive frames to get $\mathbf{M}_{\text{pst}}$ and $\mathbf{M}_{\text{fut}}$. The future root states $\mathbf{C}_{\text{fut}}$ are directly read from the motion. The model is then trained to reconstruct $\mathbf{M}_{\text{fut}}$ from a noisy version of it. For the style encoder, a \emph{different} window of length $S \in [S_{\min}, S_{\max}]$ is sampled from the \emph{same} motion sequence to yield $\mathbf{M}_{\text{syl}}$. Crucially, this dual sampling allows the style transformer to disentangle the style of the sequence from its content. We define content as the ground-projected root trajectory, while style encompasses all residual motion dynamics. The loss function combines a reconstruction loss with velocity, joint consistency, and foot sliding losses, adopted from \citep{chen_taming_2024}. 

\subsection{Diffusion Model Inference}
\label{sec:inference}
We deploy the diffusion model autoregressively, conditioning the prediction of the next $F$ frames on its own past predictions, the style $\mathbf{c}_{\text{syl}}$, and the future root states $\mathbf{C}_{\text{fut}}$. The autoregressive nature of the model, however, comes with a number of limitations, which we address through inference-time adjustments.

\paragraph{Attention Bias}
Conditioning on past motion gives the model a strong signal for how future motion should be generated, often leading it to ignore style conditioning and get stuck in a single style. To address this, we propose biasing the attention mechanism during inference. Specifically, attention in the transformer layers is computed as 
\begin{equation}
\text{attention}\left(\mathbf{Q}, \mathbf{K}, \mathbf{V}; \mathbf{B}\right) = \text{softmax}\left(\frac{\mathbf{Q} \mathbf{K}^\top}{\sqrt{D}} + \mathbf{B}\right) \mathbf{V},
\end{equation}
where $\mathbf{Q}, \mathbf{K}, \mathbf{V} \in \mathbb{R}^{N \times D}$ are the query, key and value matrices, and $\mathbf{B} \in \mathbb{R}^{N \times N}$ is the bias matrix with $N$ being the number of tokens and $D$ the model dimension. Typically, the bias matrix is set to $0$, leaving the model full control over how different tokens attend to each other. We propose adding a bias $w_{\text{syl}} \geq 0$ to the attention scores of all queries toward the style token and a bias $w_{\text{pst}} \leq 0$ to those toward the past-motion tokens. This forces the model to respect the style token, facilitating style transitions and improving stylization strength.

\paragraph{Velocity Correction}
The user controls the character through joystick commands $(\mathbf{v}_{x,y}, v_{\psi})$ comprising heading, lateral, and yaw velocities, which are integrated over timesteps $t \in \{1, \dots, F\}$ to form the trajectory conditioning signal $\mathbf{C}_{\text{fut}}$ for the diffusion model. During inference, the character can drift from the intended trajectory due to accumulation of prediction errors. Furthermore, the character is controlled through velocity commands expressed in the root heading frame. As the root naturally oscillates during movement, the generated trajectory depends heavily on where exactly in that cycle the diffusion model is triggered. 

To alleviate these problems, we keep track of the character's current position and yaw direction, $\mathbf{p}^{\text{cur}}_{x,y}$ and $\mathbf{\psi}^{\text{cur}}$, as well as the desired position and direction, $\mathbf{p}^{\text{des}}_{x,y}$ and $\mathbf{\psi}^{\text{des}}$. The current state is updated according to the diffusion model predictions whereas the desired state is computed by integrating the given joystick velocity commands $(\mathbf{v}_{x,y}, v_{\psi})$. Whenever a new diffusion call is triggered, we apply a feedback correction to the velocity command to compute corrected joystick commands:
\begin{align}
\tilde{\mathbf{v}}_{x,y} &= \mathbf{v}_{x,y} +
k_{\mathbf{p}} \mathbf{R}_\psi( \psi^{\text{cur}})^\top (\mathbf{p}^{\text{des}}_{x,y} - \mathbf{p}^{\text{cur}}_{x,y})
\\ \tilde{v}_{\psi} &= v_{\psi} + k_{\psi} (\psi^{\text{des}} - \psi^{\text{cur}}).
\end{align}
Here $k_{\mathbf{p}}$ and $k_{\psi}$ are feedback gains and $\mathbf{R}_\psi$ is the 2D rotation matrix about the yaw axis. These corrected commands $\tilde{\mathbf{v}}$ are then used to generate the future root trajectory which is used as conditioning input for the diffusion model. By treating the deviation from the intended trajectory as an error signal, this formulation introduces a closed-loop correction that continuously steers the character back toward the user's original commands. We prevent excessive corrections by projecting the desired state into a neighborhood of the current state during each timestep. Specifically, the desired position is clipped to be at most \SI{0.5}{\meter} away from the current position and the desired yaw is clipped into a \SI{0.5}{\radian} range around the current orientation. We refer to Appendix~\ref{sec:app_diffusion} for further details.

\subsection{RL-Based Tracking Controller}
\label{sec:RL}
We use RL to train a policy that tracks the generated kinematic motions, largely following related work \cite{serifi_vmp_2024, peng_deepmimic_2018}. The policy predicts actions given observations and goal states. We refer to Appendix~\ref{sec:app_rl} for details on the design of the action space, rewards, observations, and goal states.

\paragraph{Training Data} We avoid the computational overhead of querying the diffusion model during RL by generating a large motion dataset offline via autoregressive rollouts. To do so, we run the diffusion model conditioned on style clips randomly sampled from the diffusion model's training data paired with randomly sampled joystick commands. This ensures that we cover the whole space of motions that can be generated by the diffusion model. 

\paragraph{Deployment}
We deploy the policy and diffusion model in an open loop: the diffusion model is conditioned on its own past predictions, rather than the real state of the character, which we found to improve stylization quality. Furthermore, an open-loop scheme allows training on a precomputed dataset instead of running the diffusion model online. This is feasible because we do not deal with precise object or environment interactions that require the diffusion model to adjust to the real-world data, as in other works \citep{tevet_closd_2024, zhang2026learningwholebodyhumanoidlocomotion}. To avoid computational overhead from the style encoder, we precompute a set of style encodings and cache them during deployment.

\section{Experiments and Results} \label{sec:results}

We first compare our method to CAMDM~\cite{chen_taming_2024} and analyze our style encoder. We then evaluate the attention biasing and velocity correction. Finally, we ablate our RL tracking policy and transfer our method onto a real robot. We refer to the Appendix for implementation details.

\subsection{Diffusion Model}
\label{sec:exp_diffusion}
\paragraph{Quantitative Evaluation}
We compare our diffusion model against CAMDM \cite{chen_taming_2024}, the current state of the art in generative modeling of stylized motion and the only model directly applicable to our setting when evaluating seen styles (\Tab{comparison}). Both models are trained on the 100STYLE dataset \cite{mason_2022_real}. We report FID score~\cite{heusel2017gans}, style accuracy, foot sliding, and positional and yaw errors with respect to the condition trajectory (see Appendix \ref{sec:app_eval}). We summarize results in \Tab{diffusion_quality} and provide a qualitative comparison in Appendix~\ref{sec:app_qual}. Compared to CAMDM, our model generates motion that is closer to the real motion capture data, as indicated by the lower FID and higher style accuracy, while achieving similar results in terms of physicality and alignment to the trajectory conditions.  Furthermore, we find that adding attention biasing at inference improves overall motion quality for both CAMDM and our model, which we discuss in more detail in \Sec{exp_inference}.

\input{sections/tables/diffusion_quality}

\paragraph{Style Encoder}
\begin{figure}
    \centering
    \includegraphics{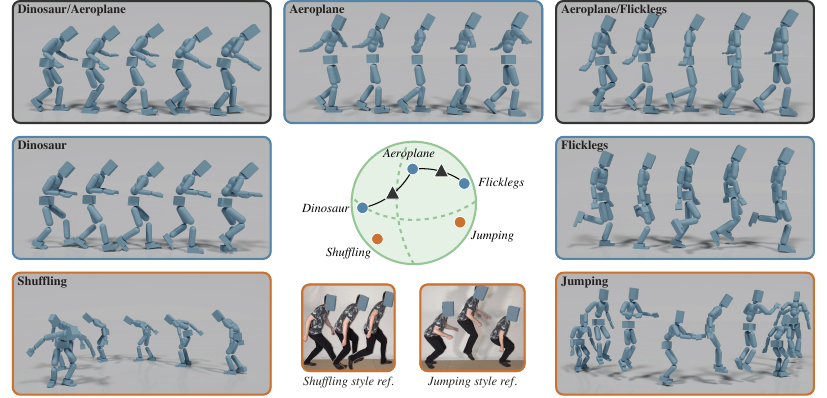}
    \caption{\textbf{Style Encoding.} Generated motions conditioned on interpolated style latents of styles from the 100STYLE dataset (top and middle row) and latents computed from unseen motions extracted from videos (bottom row).}
    \label{fig:style_enc_examples}
\end{figure}

We highlight the interpolation and generalization capabilities of our style encoder.  First, we encode styles from the 100STYLE dataset and spherically interpolate between the corresponding style latents. Next, we evaluate the generalization of our method by capturing videos of a person performing a given style. We then extract 120 frames of motion features from that video using an open-source pose estimation model \citep{shen2024gvhmr} and pass them as input to the style encoder. Notably, no further processing after pose estimation is needed before feeding the clips to the style encoder. As seen in \Fig{style_enc_examples} and our supplemental video, our model interpolates smoothly between styles and even generalizes to a highly dynamic style such as jumping. For this experiment, we train the model on both the 100STYLE \citep{mason_2022_real} and AMASS \citep{AMASS:ICCV:2019} datasets to improve generalization capabilities. Furthermore, \Tab{diffusion_quality} shows quantitative evaluations on a set of 10 styles of 100STYLE not included during training. Importantly, CAMDM is not applicable here, as it is inherently limited to seen styles. On unseen styles, our model reproduces their distinctive features and reaches style accuracy comparable to that on seen styles, but it struggles with finer details, leading to a higher FID. Visualizations of the motions generated for the test split can be found in Appendix~\ref{sec:app_qual}. Finally, we apply three additional probes to the style latents for further evidence of disentanglement between style and content (the ground-projected root trajectory). First, linear discriminant analysis \citep{fisher_lda} recovers style identity from the style latents with $95\%$ accuracy across all styles of 100STYLE. Second, within-style variation of the style latents, computed across different clips of the same style, accounts for only $19\%$ of the total style latent variance. Third, regressing content from the style latents alone yields $R^2 = 0.07$, barely better than predicting the mean content.

\paragraph{Style Clip Sampling Strategy} 
We verify the importance of our style encoder sampling strategy by training a model where we always sample the \emph{same} motion window for both style reference and past and future motion. Using the \emph{same} motion window results in a sample diversity of \SI{0.8}{\deg} for joint angles across different generated motion windows, while using \emph{different} motion windows (our approach) results in \SI{5.5}{\deg}.
We refer to Appendix \ref{sec:app_qual} for qualitative results, which show that without our sampling scheme, the model collapses, causing all generated motion windows to repeat the reference motion. In contrast, with our sampling scheme, we can disentangle the style from the locomotion and generate diverse motion. 

\subsection{Inference Techniques}
\label{sec:exp_inference}
\paragraph{Attention Bias}
\label{sec:results_attention_bias}
We compare our proposed attention bias method to CFG~\cite{ho2022classifierfreediffusionguidance} on past motion where the predicted motion is a weighted average of a fully conditioned and a masked prediction (see Appendix \ref{sec:app_eval}). 
\Tab{diffusion_quality} shows how attention biasing significantly improves quality and stylization of generated motions. As it is a pure inference adjustment, we can apply it to both models, \textit{Ours + Bias} and \textit{CAMDM + Bias}, though the relative improvement is smaller for CAMDM because its discrete style labels are less informative than our style encodings. Improved stylization comes at the cost of more foot sliding, as negative weighting of past motion makes it more challenging to create coherent foot swings. Compared to CFG, attention biasing performs better in all metrics. We conjecture that this performance gap arises because the bias modifies the pre-softmax weights, so attention stays a normalized distribution over the same tokens and thus closer to the training regime. In contrast, CFG interpolates two separate in-distribution predictions at every denoising step, and the result of this guidance step may no longer lie in the training distribution. Qualitative results and a sensitivity analysis with respect to the bias weights can be found in the Appendix (\Fig{attention_bias} and \Fig{bias_sweep} respectively).

\clearpage
\newsavebox{\attentionbiasbox} %
\sbox{\attentionbiasbox}{%
    
    \begin{tabular}{lcc}
    \toprule
    \textbf{Method} & Succ. Rate $\uparrow$ & Inf. Time $\downarrow$ \\
    \midrule
    None            & $83.8\%$                     & \SI{7.69}{\milli\second}                     \\
    CFG             & $98.7\%$                     & \SI{14.30}{\milli\second}                     \\
    Bias            & $98.6\%$                     & \SI{8.50}{\milli\second}                      \\
    \bottomrule
    \end{tabular}%
}

\begin{wraptable}[5]{r}{\wd\attentionbiasbox}
    \centering
    \label{tab:transitions}

    \renewcommand{\arraystretch}{1.0}
    \setlength{\aboverulesep}{2pt}
    \setlength{\belowrulesep}{2pt}
    \setlength{\cmidrulesep}{1pt}
    
    \usebox{\attentionbiasbox} %
\end{wraptable}
We also investigate style transition behavior between the different methods more closely. As shown in the inset table, both our attention biasing and CFG achieve similar improvements over regular diffusion rollouts in transition success rates \cite{chen_taming_2024}. However, our method incurs much less computational overhead than CFG, which requires two model evaluations in each diffusion step.

\paragraph{Velocity Correction}

\sbox{\mytablebox}{%
  \begin{tabular}{ccc}
  \toprule
  \shortstack{\textbf{Velocity} \\ \textbf{Correction}} & \shortstack{Drift\\per meter  $\downarrow$} & \shortstack{Mean \\ vel. error $\downarrow$} \\
  \midrule
  \xmark & $0.117$ & \SI{0.271}{\meter\per\second} \\
  \cmark & ${0.003}$ & \SI{0.075}{\meter\per\second} \\
  \bottomrule
  \end{tabular}%
}
\begin{wraptable}[5]{r}{\wd\mytablebox}%
\vspace{-2.5ex} 
\centering

\label{tab:velocity_correction_selected_split}
\usebox{\mytablebox} %
\end{wraptable}%
We evaluate how closely the generated motion adheres to the intended trajectories, with and without velocity correction. To that end, we deploy the diffusion model with velocity commands corresponding to a number of predefined trajectories, such as squares, circles, figure eights, and randomized velocity commands. The quantitative results, reported in the inset table, highlight that our correction scheme significantly improves adherence to the given commands, indicated by the reduced drift and mean velocity error. We refer to Appendix \ref{sec:app_qual} and \Fig{velcorrect_qual} for qualitative results and details on the test trajectories.
\subsection{Physics-Based Tracking}
\paragraph{RL Tracking Policy}

\begin{table}[]
    \centering
    \caption{\textbf{RL Tracking Policy.} We evaluate the effect of training data and loop mode on the performance of the tracking policy. Additionally, we compare open- and closed-loop performance under strong external disturbances.}
    \begin{tabular}{llccccc} 
         \toprule
        \textbf{Training Data} & \textbf{Loop Mode} & \textbf{Disturbances} & \multirow{2}{*}{FID\,$\downarrow$}
        & Style Acc. & MPJAE & Falls\\
        & & & & [\%] $\uparrow$ & [\unit{\deg}] $\downarrow$ & [\#/\si{\minute}] $\downarrow$ \\
        \midrule
        MoCap & Open & \xmark & 0.16 & 92.0 & 5.2 & 1.104\\
        Generated & Open & \xmark & 0.35 & 85.1 & 8.3 & 0.001\\
        Generated & Closed & \xmark & 0.45 & 79.6 & 7.8 & 0.001\\
        \midrule
        Generated & Open & \cmark & 0.84 & 62.9 & 16.0 & 3.812\\
        Generated & Closed & \cmark & 0.95 & 58.2 & 15.9 & 4.431\\
        \bottomrule
    \end{tabular}
    \label{tab:tracking_eval}
\end{table}

\begin{figure}
    \includegraphics[width=\linewidth]{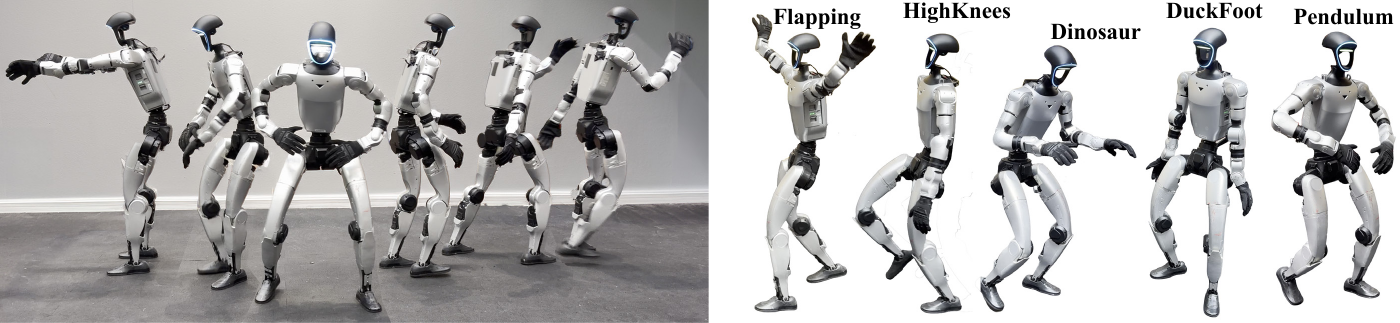}
    \caption{\textbf{G1 Robot Deployment.} Left: Continuous joystick control with dynamic style modulation. Right: Demonstration of the framework generating five distinct motion styles.}
    \label{fig:g1}
    \vspace{-2mm}
\end{figure}

We evaluate how the choice of training data affects the RL policy's tracking of diffusion-generated motions. \Tab{tracking_eval} shows that training a policy on the original, non-diffusion-generated data has a significantly higher failure rate (falls per minute) when tracking diffusion output than when training on generated data. This happens because the policy trained on the diffusion output learns to correct artifacts, such as foot sliding, yielding improved stability at the cost of higher tracking errors and worse motion quality. We further compare our open-loop approach, where the diffusion model is conditioned on its own past predictions, with a closed-loop approach, where the diffusion model is conditioned on states estimated from the robot. The open-loop approach leads to better stylization since, in closed loop, the past simulated states differ in distribution from the past motions the diffusion model was trained on. Closed-loop execution leads to lower tracking errors, because the tracking errors enter the diffusion conditioning, pulling the generated motion toward what the policy executes rather than toward the intended style. Finally, we apply perturbation forces of \SI{600}{\newton} to the character root, resulting in lower motion quality, higher tracking errors, and significantly more failures for both open and closed loop. Open loop nevertheless remains more stable than closed loop. Recovery states can enter the diffusion conditioning in closed-loop mode, pushing the diffusion model out of distribution. In contrast, with an open-loop deployment the diffusion model's generation stream stays uncorrupted and provides clean recovery targets for the policy. We found no degradation of motion quality over the whole time horizon of 20 minutes per rollout. The tracking evaluations discussed so far were performed for styles contained in the training set of both the diffusion model and RL policy. For styles that are generated by interpolating between style latents of training styles, as seen in \Fig{style_enc_examples}, tracking performance stays consistent at an MPJAE of \SI{7.9}{\deg} and 0 falls. FID and style accuracy cannot be computed for such interpolated styles, as no ground-truth data exists.

\paragraph{Transfer to Physical Robot}
We apply our method to a Unitree G1 humanoid robot (\SI{1.32}{\meter}, \SI{35}{\kilo\gram}, $29$ DoF). After retargeting the human data to the G1 with ReActor~\citep{muller2026reactor}, we train our model and deploy it zero-shot on the robot. Qualitative results, provided in \Fig{g1} and our supplemental video, show that a single controller can robustly track diverse motion styles using different joystick inputs. The robustness of the open-loop mode carries over to hardware, including under disturbances. Throughout $\sim$\SI{4}{\hour} of deployments, the robot successfully executed various trajectories and velocity commands across different styles without any failures. The diffusion model runs in real time on an onboard NVIDIA Jetson Thor at $\sim$\SI{11}{\milli\second} per inference step and $\sim$\SI{17}{\milli\second} command-to-motion output latency, compared to \SI{8}{\milli\second}/\SI{12}{\milli\second} for simulation on an NVIDIA RTX~5090.

\section{Limitations}
While our work presents a framework for general, stylized locomotion control, several limitations remain. The RL policy can correct artifacts of the diffusion model (foot sliding, hovering, ground penetration). However, there can be a mismatch between the kinematic motions and the commanded velocity, which manifests as the tracking policy taking extra in-between steps to maintain stability. Introducing implicit rewards to achieve natural gaits could present a solution to these issues \citep{mu2026smp, peng_amp_2021}. Our method works with styles that are agnostic to the direction of movement, such as flailing the arms or walking in a wide stance. For styles that are tightly coupled to the direction of movement, such as a moonwalk or cartwheels, our method requires matching velocity commands. Furthermore, styles that are largely described through a distinct root trajectory, such as rotating clockwise while walking, specify the same quantity as the joystick, creating an inherent conflict in the interface, which we resolve in favor of the user command. Finally, our current framework lets the model learn the style implicitly. A more explicit parameterization could allow users to partially stylize, for example, focusing only on the lower or upper body behavior of a style reference.

\section{Conclusion}
We presented a real-time system for expressive legged locomotion that enables simultaneous user control over velocity and style via an example-driven, autoregressive diffusion model. Through a style encoder that is trained with an effective sampling strategy on short, unstructured motion clips rather than fixed style labels, we enable physics-based characters to generalize to completely unseen styles, including those extracted from real-world videos. This flexible interface opens the door to various other style input modalities, such as artist-created animations.

\clearpage
\acknowledgments{We would like to thank Ioannis Dadiotis for his support with robot deployments.}

\bibliography{example}  %

\clearpage
\appendix
\counterwithin{figure}{section}
\counterwithin{table}{section}
\input{sections/appendix}

\end{document}

%% file: sections/tables/diffusion_quality.tex
\begin{table}[t]
    \centering
    \caption{\textbf{Quantitative Evaluation.} We compare our method to CAMDM \cite{chen_taming_2024} and evaluate the effects of attention biasing and CFG.}
    \begin{adjustbox}{width=\columnwidth,center}
    \scriptsize 
    \renewcommand{\arraystretch}{1.0} 
    
    \begin{tabular}{l ccccc}
        \toprule
        \multirow{2}{*}{\textbf{Method}} & \multirow{2}{*}{FID $\downarrow$} & Style Acc. $\uparrow$ & Ft. Slid. $\downarrow$ & Pos. Err. $\downarrow$ & Yaw Err. $\downarrow$ \\
                                                                            &                                   & [\%]                  & [\unit{\centi\meter}/frame]             & [\unit{\centi\meter}/frame]             & [\unit{\deg}/frame]           \\
        \midrule
         
        CAMDM \cite{chen_taming_2024}        & 0.27 & 86.0 & 0.71 & 0.70 & 0.60 \\
        Ours           & 0.22 & 88.1 & \textbf{0.66} & 0.70 & 0.65 \\ \midrule
        CAMDM \cite{chen_taming_2024} + Bias  & 0.14 & 93.5 & 0.90 & \textbf{0.69} & \textbf{0.58} \\
        Ours + CFG    & 0.17 & 91.5 & 0.98 & 0.71 & 0.68 \\
        Ours + Bias   & \textbf{0.07} & \textbf{96.3} & 0.93 & 0.71 & 0.67 \\ 
        \midrule 
        Ours, Unseen Styles & 0.34 & 93.0 & 0.90 & 0.71  & 0.69 \\
        \bottomrule
    \end{tabular}
    \end{adjustbox}
    \label{tab:diffusion_quality}
    \vspace{-2mm}
\end{table}

%% file: sections/appendix.tex
\setlength{\textfloatsep}{12pt plus 2pt minus 2pt}  %
\setlength{\floatsep}{8pt plus 2pt minus 2pt}       %

\section{Appendix}

\subsection{Kinematic Diffusion Model}
\label{sec:app_diffusion}

\paragraph{Motion Representation} 
We represent the motion state, similarly to CAMDM~\cite{chen_taming_2024}, as
\begin{equation}
\mathbf{m} = ( \mathbf{x}, \mathbf{r}, \mathbf{J}, \mathbf{X}, \mathbf{R} ),
\label{eq:motion_repr}
\end{equation}
with the position $\mathbf{x} \in \mathbb{R}^3$ and orientation $\mathbf{r} \in \mathbb{R}^6$ of the character's root, the orientations of all its $N_J$ joints $\mathbf{J} \in \mathbb{R}^{N_J \times 6}$, and the positions $\mathbf{X} \in \mathbb{R}^{N_B \times 3}$ and orientations $\mathbf{R} \in \mathbb{R}^{N_B \times 6}$ of all its $N_B$ links. We represent all rotations in 6D \cite{zhou_continuity_2019}. For joint orientations, we represent a child's orientation in the kinematic chain by its orientation in the parent's coordinate frame. 

To train a model that is agnostic to the character's global state, we transform all quantities, except the joint orientations, in $\mathbf{M}_{\text{pst}}$ and $\mathbf{M}_{\text{fut}}$, to the heading frame of the character, extracted from the current state $\mathbf{m}_t$. The origin of this frame is set to the character's ground-projected root position, its x-axis points in the direction of the root's yaw, and the z-axis is aligned with the global up vector.

\paragraph{Model and Training Parameters}
\begin{wraptable}[7]{right}{4cm}
    \vspace{-2.7ex}
    \centering
    \small
    \begin{tabular}{lr}
         \toprule
         Parameter & Value \\
         \midrule
         $d_{\text{model}}$& 256 \\
         FeedForward Size & 1024 \\
         \# Layers & 4 \\
         \# Heads & 4 \\
         Dropout Rate & 0.2\\
         Activation Fn. & GELU \\
         \bottomrule
    \end{tabular}
\end{wraptable}
Both the diffusion model $\mathcal{M}$ and style encoder $\mathcal{E}$ are realized as transformer-encoders \citep{vaswani_attention_2017} with the same model parameters which can be seen in the inset table. These parameters are used across all experiments for both the humanoid and robotic characters. We train all models with 8 diffusion steps with a cosine noise schedule. Training is performed for 100 epochs and the model weights are optimized using Adam \citep{kingma2017adammethodstochasticoptimization} with default parameters except for the learning rate, which is set to $10^{-4}$. During training we randomly mask past motion by setting it to $0$ with probability $0.15$. For the humanoid character we represent motion at $30$ frames per second and set the future frames $F = 20$ with past frames $P = 10$. The robotic character uses $50$ frames per second with $F=30$ and $P=15$. To ensure responsiveness to velocity commands, we trigger early autoregression after 10 and 15 frames, respectively, if the velocity commands change. For both characters, we sample style clips of lengths \SI{1}{\second} to \SI{5}{\second}. An important consideration is that the humanoid uses spherical joints while the robot uses revolute joints. For the diffusion model itself, we treat both joint types as if they were spherical joints, which allows the diffusion model to predict infeasible orientations for the revolute joints. We then post-process the diffusion model predictions by projecting the predicted orientations onto the set of feasible orientations using a swing-twist decomposition \citep{dobrowolski2015swingtwistdecompositioncliffordalgebra}.

\paragraph{Motion Smoothing} To alleviate discontinuities between consecutive diffusion predictions, we apply a smoothing scheme to generate motion. Specifically, rather than generating a new motion every $F$ frames, we trigger the diffusion model after only $F-K$ frames, creating a $K$-frame overlap between consecutive predictions, as visualized in \Fig{smoothing}. Then, the overlapping frames are blended together, using linear and spherical interpolation for positions and rotations, respectively. In practice, we set $K = 3$.

\paragraph{Velocity Correction} 
For the feedback gains we set $k_{\mathbf{p}} = k_{\psi} = \frac{1}{F \Delta t}$ where $F$ is the number of future frames predicted by the diffusion model and $\Delta t$ is the duration of each timestep, i.e., the inverse frame rate. This ensures that the error is corrected over the span of one diffusion prediction, without under-correcting or overshooting. Finally, we use the mean yaw error over the last $P$ frames when computing the feedback update. This prevents the error correction itself from depending on when exactly in the gait cycle diffusion is triggered.

\clearpage
\input{sections/figures/smoothing}

\subsection{Reinforcement Learning}
\label{sec:app_rl}

We use RL to train a policy $\pi(\mathbf{a}_t \,|\, \mathbf{o}_t, \mathbf{g}_t)$ to track the generated motion. The policy predicts joint setpoints $\mathbf{a}_t$ for proportional-derivative (PD) controllers based on observations $\mathbf{o}_t$ and goal conditioning $\mathbf{g}_t$.
The conditioning $\mathbf{g}_t$ contains the character's target state at timestep $t$ as predicted by the diffusion model, converted into the character's heading frame of step $t-1$. Additionally, we explicitly concatenate the joystick command to the target state. Proprioceptive observations consist of joint positions and velocities, root angular velocity, projected gravity and a history of previous actions.

\paragraph{Observations and Goals}
In our tracking policy, we use the goal condition 
\begin{equation}
    \mathbf{g}_t \coloneq ( \mathbf{x}_t, \mathbf{r}_t, \mathbf{q}_t, \mathbf{X}_t, \mathbf{R}_t,  \dot{\mathbf{x}}_t, \bm{\omega}_t, \tilde{\mathbf{v}}_t), 
\end{equation}
containing the set of reference motion features $\mathbf{m}_{t}$ (see \Eq{motion_repr}). However, note that here we represent the joint angles in minimal coordinates $\mathbf{q}_t$ instead of using $\mathbf{J}_t$, as is common for the simulation of articulated bodies. For the robot with revolute joints, $\mathbf{J}$ is projected to a feasible $\mathbf{q}_t$. In addition, $\mathbf{g}_t$ includes the target root linear velocity $\dot{\mathbf{x}}_t$ and angular velocity $\bm{\omega}_t$, computed from $\mathbf{M}_{\text{fut}}$, as well as the velocity command after correction $\tilde{\mathbf{v}}_t$ (\Sec{inference}). 
All rigid body positions and orientations are expressed in the root's heading frame, making the policy independent of the global state. 
The proprioceptive observations $\mathbf{o}_t$ are
\begin{equation}
    \mathbf{o}_t \coloneq (\hat{\mathbf{q}}_t, \hat{\dot{\mathbf{q}}}_t, \hat{\bm{\omega}}_t, \hat{\mathbf{g}}^{\text{proj}}_t, \mathbf{a}_{t-1}, \mathbf{a}_{t-2}),
\end{equation}
where the hat symbol ($\hat{\cdot}$) indicates measured values from the simulator or the real system. These observations include the projected gravity $\hat{\mathbf{g}}^{\text{proj}}_t$, angular velocity $\hat{\bm{\omega}}_t$, joint angles $\hat{\mathbf{q}}_t$, joint velocities $\hat{\dot{\mathbf{q}}}_t$, as well as the previous two actions $\mathbf{a}_{t-1}$ and $\mathbf{a}_{t-2}$.

\paragraph{Rewards}
We use a combination of three groups of reward functions, consisting of tracking rewards $\mathbf{r}_{\text{track}}$, regularization rewards $\mathbf{r}_{\text{reg}}$, and task rewards $\mathbf{r}_{\text{task}}$. Tracking rewards ensure that the policy follows the prescribed motion, while the regularization rewards ensure that it does so in a smooth and natural manner that is deployable on real hardware. The task reward incentivizes following the joystick commands and helps with resolving unphysical target states, for example when the diffusion model generates motion with foot sliding. In such cases, the task reward encourages the policy to give less priority to exact tracking, resulting in more natural and stable motion. \Tab{rewards} shows all the individual reward terms with their respective weights.

\begin{table}[]
  \centering
      \caption{\textbf{RL Reward Terms.} Reward terms and weights for the Humanoid and the G1 robot. Tracking
  and joystick rewards use an exponential kernel $\exp(-e/\sigma)$ on the listed squared error $e$. We drop the time index $t$ for conciseness. Norms are Euclidean unless specified otherwise. $h_f$ measures the foot-ground distance while $c_f^\text{sim},c_f^\text{ref}$ are simulation or reference provided contact flags.}
      \label{tab:rewards}
      \begin{tabular}{llccc}
      \toprule
      \textbf{Name} & \textbf{Error / Term} & $\sigma$ & \textbf{Humanoid} & \textbf{G1 Robot} \\
      \midrule
      \midrule
      \multicolumn{5}{c}{\textit{Tracking}} \\
      \midrule
      Root position & $\|\mathbf{x} - \hat{\mathbf{x}}\|^2$ & 0.1 & 1.0 & 2.0 \\
      Root orientation & $\|\mathbf{r} \ominus \hat{\mathbf{r}}\|_F^2$ & 0.5 & 1.0 & 2.0 \\
      Root lin. velocity & $\|\dot{\mathbf{x}} - \hat{\dot{\mathbf{x}}}\|^2$ & 1.0 & 1.0 & 2.0 \\
      Root ang. velocity & $\|\bm{\omega} - \hat{\bm{\omega}}\|^2$ & 1.0 & 1.0 & 2.0 \\
      Upper body joints & $\|\mathbf{q}^u - \hat{\mathbf{q}}^u\|^2$ & 0.5 & 5.0 & 5.0 \\
      Lower body joints & $\|\mathbf{q}^l - \hat{\mathbf{q}}^l\|^2$ & 0.5 & 1.0 & 3.0 \\
      Ankle joints & $\|\mathbf{q}^a - \hat{\mathbf{q}}^a\|^2$ & 1.0 & 1.0 & 1.0 \\
      Body positions & $\|\mathbf{X} - \hat{\mathbf{X}}\|^2$ & 1.0 & 1.0 & 1.0 \\
      Body orientations & $\|\mathbf{R} \ominus \hat{\mathbf{R}}\|_F^2$ & 2.5 & 1.0 & 1.0 \\
      Body lin. velocity & $\|\dot{\mathbf{X}} - \hat{\dot{\mathbf{X}}}\|^2$ & 5.0 & -- & 2.0 \\
      Alive bonus & $1$ & -- & 1.0 & 2.0 \\
      \midrule
      \multicolumn{5}{c}{\textit{Joystick commands}} \\
      \midrule
      Joystick lin. velocity & $\|\tilde{\mathbf{v}}_{x,y} - \hat{\mathbf{v}}_{x,y}\|^2$ & 0.25 & 5.0 & 3.0 \\
      Joystick yaw rate & $(\tilde{v}_{\psi} - \hat{v}_{\psi})^2$ & 0.25 & 5.0 & 5.0 \\
      Lin. velocity direction & $\cos\angle(\tilde{\mathbf{v}}_{x,y}, \hat{\mathbf{v}}_{x,y}) \cdot \mathbbm{1}[\|\tilde{\mathbf{v}}_{x,y}\| > 0.2]$ & -- & -- & 5.0 \\
      Yaw rate direction & $\operatorname{sgn}(\tilde{v}_{\psi})\operatorname{sgn}(\hat{v}_{\psi}) \cdot \mathbbm{1}[\lvert \tilde{v}_{\psi} \rvert > 0.2]$ & -- & -- & 5.0 \\
      Stuck (lin. velocity) & $\mathbbm{1}[\|\hat{\mathbf{v}}_{x,y}\| < 0.2 \,\wedge\, \|\tilde{\mathbf{v}}_{x,y}\| > 0.2]$ & -- & -- & $-10.0$ \\
      Stuck (yaw rate) & $\mathbbm{1}[\lvert \hat{v}_{\psi} \rvert < 0.2 \,\wedge\, \lvert \tilde{v}_{\psi} \rvert > 0.2]$ & -- & -- & $-10.0$ \\
      \midrule
      \multicolumn{5}{c}{\textit{Regularization}} \\
      \midrule
      Action rate & $\|\mathbf{a}_t - \mathbf{a}_{t-1}\|^2$ & -- & $-0.01$ & $-0.1$ \\
      Action accel. & $\|\mathbf{a}_t - 2\mathbf{a}_{t-1} + \mathbf{a}_{t-2}\|^2$ & -- & $-0.01$ & $-0.05$ \\
      Joint accel. & $\|\ddot{\mathbf{q}}\|^2$ & -- & $-10^{-5}$ & $-10^{-6}$ \\
      Torque & $\|\bm{\tau}\|^2$ & -- & $-10^{-5}$ & $-10^{-5}$ \\
      Joint position limits & $\sum_j \max(0,\, \lvert q_j \rvert - q_j^{\max})$ & -- & -- & $-10.0$ \\
      Joint velocity limits & $\sum_j \max(0,\, \lvert \dot{q}_j \rvert - 0.95\, \dot{q}_j^{\max})$ & -- & -- & $-1.0$ \\
       Foot impact & $\sum_{f} \max(0,\, \|\mathbf{F}_{f}\| - 400)$ & -- & -- & $-0.01$ \\
      Simulation-contact slip & $\sum_{f} \|\hat{\dot{\mathbf{X}}}_{f,xy}\| \cdot c_f^\text{sim}$ & -- & -- & $-0.3$ \\
      Reference-contact slip & $\sum_{f} \|\hat{\dot{\mathbf{X}}}_{f,xy}\| \cdot {c}_{f}^\text{ref}$ & -- & -- & $-5.0$ \\
      Swing clearance & $\sum_{f} (1 - {c}_{f}^\text{ref}) \cdot \max(0,\, 0.02 - h_{f})$ & -- & -- & $-200.0$ \\
      \bottomrule
      \vspace{-20pt}
      \end{tabular}
\end{table}

\paragraph{Training Environment} 

We follow standard practices in the training of tracking controllers \citep{serifi_vmp_2024, peng_deepmimic_2018}. During each episode we sample a random motion from the training set and iterate through it to get the individual goals of each timestep. Here, we also randomly sample the starting frame inside the goal motion and combine it with reference state initialization where the character is initialized in that exact state \citep{peng_deepmimic_2018}. To allow the policy to reach any target state from any 
\begin{wraptable}[10]{right}{6.5cm}
    \vspace{-10pt}
    \centering
    \small
    \begin{tabular}{lr}
        \toprule
         Parameter & Value \\
         \midrule
         $\varepsilon_{\text{clip}}$ & 0.2\\
         Value Loss Coefficient & 1.0 \\
         Entropy Coefficient & 0.005  \\
         Discount Factor $\gamma$ & 0.99 \\
         GAE Factor $\lambda$ & 0.99\\
         Desired KL Divergence & 0.01\\
         \# Environments & 4096\\
         Env. Steps per PPO Iter. & 32 \\
         Learning Epochs per PPO Iter. & 5\\
         \# Mini Batches & 4 \\
         Schedule & Adaptive \\
        \bottomrule
    \end{tabular}
\end{wraptable}
starting state we choose the reference state during initialization from a different motion than the actual target motion with a probability of $0.2$. Episodes terminate early if the pelvis or head contacts the ground. The policy itself is trained with PPO \citep{schulman_proximal_2017} with the hyperparameters given in the inset table.

\paragraph{Training Data} To generate the training data for the policy we generate a large-scale dataset using diffusion model rollouts with randomized commands. For each clip of motion in the training set of the diffusion model, we generate 10 minutes of motion. To do so, we randomly resample velocity commands every \SI{1.5}{\second} to \SI{4.5}{\second} and randomly resample the chosen style reference window from the clip every \SI{5}{\second}. Additionally, we generate another set of 10 hours of motion where we sample velocity commands in the same manner while also randomly sampling the style reference over the whole training set, to ensure that the policy can transition between different styles. 

\paragraph{Robustness and Sim-to-Real} In order to improve the robustness of the policies and ensure feasibility of transferring them to the real world we apply disturbances and domain randomization \cite{Sim2Real2018} with their respective values shown in \Tab{domainrand}. A curriculum is applied to the magnitude of disturbances, which ramps linearly from 0 to the specified maximum values between PPO iterations 300 and 2000. Finally, the policies are trained on rough terrain where height field values in the range $[-2, 2]$ \SI{}{\centi\meter} are uniformly sampled and placed on a grid with spacing of \SI{10}{\centi\meter}. The grid points are then linearly connected to generate the ground terrain. For the G1 robot, we use the same actuator parameters as in \citep{wang2026omnixtremebreakinggeneralitybarrier}.

\begin{table}[]
    \caption{\textbf{Domain Randomization.} Randomization ranges applied during policy training. A dash indicates that the term is not randomized for the respective embodiment.}
        \label{tab:domainrand}
    \centering
    \begin{tabular}{lccc}
    \toprule 
    \textbf{Term} & \textbf{Unit} & \textbf{Humanoid} & \textbf{Robot} \\
    \midrule
    \midrule
    \multicolumn{4}{c}{\textit{Additive Observation Noise}} \\
    \midrule
    Projected Gravity & \SI{}{\radian} & $-$ & $\mathcal{U}(-0.05, 0.05)$ \\
    Root Angular Velocity & \SI{}{\radian\per\second} & $-$ & $\mathcal{U}(-0.2, 0.2)$ \\
    Joint Position & \SI{}{\radian} & $-$ & $\mathcal{U}(-0.01, 0.01)$ \\
    Joint Velocity & \SI{}{\radian\per\second} & $-$ & $\mathcal{U}(-1.5, 1.5)$ \\
    \midrule
    \multicolumn{4}{c}{\textit{Physics Randomization}} \\
    \midrule
    Root Mass Offset & \SI{}{\kilo\gram} & $-$ & $\mathcal{U}(-1.0, 3.0)$ \\
    Root/Link CoM Offset & \SI{}{\meter} & $-$ & $\mathcal{U}(-0.05, 0.05)$ \\
    Foot Friction & $-$ & $\mathcal{U}(0.4, 0.6)$ & $\mathcal{U}(0.2, 1.0)$ \\
    $K_p$, $K_d$ scale & $-$ & $-$ & $\mathcal{U}(0.9, 1.1)$ \\
    \midrule
    \multicolumn{4}{c}{\textit{Disturbances}} \\
    \midrule
    Push Root & \SI{}{\meter\per\second} & $\mathcal{U}(-0.5, 0.5)$ & $\mathcal{U}(-0.5, 0.5)$ \\
    Push Ankle Force & \SI{}{\newton} & $5.0$ & $5.0$ \\
    Push Ankle Torque & \SI{}{\newton\meter} & $2.5$ & $2.5$ \\
    \bottomrule
    \end{tabular}
    \vspace{-3mm}
\end{table}

\subsection{Evaluations}
\label{sec:app_eval}
\paragraph{Evaluation Setup}
We train the diffusion model for 100 epochs. When training on 100STYLE \citep{mason_2022_real} this corresponds to roughly two days of training on an NVIDIA RTX 4090. With the inclusion of AMASS \citep{AMASS:ICCV:2019}, training time increases to roughly five days. To ensure convergence, the RL policies are trained for five days on an NVIDIA RTX 4090. However, reasonable performance is already achieved after a few hours of training. For all experiments involving attention biasing we set $w_{\text{syl}}=2.0, w_{\text{pst}}=-1.0$ and for CFG we set $\gamma_{\text{CFG}}=0.6$. To assess the diffusion model quality, we generate 20 minutes of motion for each of the 100 styles in 100STYLE. For the transitions, we generate 5 transition motions from each style to every other style in 100STYLE. Here, we first generate \SI{5}{\second} of motion in style $i$, then switch the style reference to style $j$ and generate another \SI{5}{\second} of motion. Velocity commands are randomly sampled for both the diffusion model and transition evaluations. For the evaluations of the RL policies we generate and track 20 hours of motions with randomized commands.

\paragraph{Evaluation Metrics}
For the evaluation of the diffusion model we rely on a number of metrics to assess motion quality, adherence to commands as well as physicality of the motion. First of all, we use the Fréchet Inception Distance (FID)~\cite{heusel2017gans} as a measure of distributional similarity between original MoCap and generated motions. The FID value is computed as 
\begin{equation}
    \text{FID} = \|\bm{\mu}_{r} - \bm{\mu}_g\|_2^2 + \text{Tr}\left(\bm{\Sigma}_r + \bm{\Sigma}_g - 2(\bm{\Sigma}_r \bm{\Sigma}_g)^{\frac{1}{2}}\right)
\end{equation}
    
where $\bm{\mu}_r, \bm{\mu}_g$ and $\bm{\Sigma}_r, \bm{\Sigma}_g$ are empirical means and covariances of latent features of the real and generated data respectively. We extract these latent features from the activations of a classification model. This classification model is trained to classify \SI{0.25}{\second}  motion windows of the original 100STYLE dataset into its corresponding style and we use the final activations, prior to the classification head of the model, as latent representations. Importantly, FID score is computed on a per-style basis so that it captures stylistic features of motion rather than just generic similarity to the MoCap clips. Style accuracy is computed as the percentage of generated motions for which the style classifier predicts the style label that corresponds to the motion that was used as style reference. Position and yaw errors are simply computed as the difference between predicted position/yaw and their imposed condition counterparts. We compute foot sliding by measuring foot movement under estimated contact where a foot is classified as in contact if its heel height is less than \SI{1}{\centi\meter}. For transitions we compute the success rate based on the style classifier: when the model should transition from style $i$ to style $j$, we count it as a success if the style classifier predicts label $j$ after the transition is initiated. We do not impose a maximum transition time, so a transition counts as successful even if style $j$ is only reached by the end of the predicted motion.

\paragraph{CFG}
In our experiments, we compare against CFG on past motion, where each diffusion step is computed as 
\begin{equation}
\begin{split}
\mathbf{M}^0_{\text{fut}} &= \gamma_{\text{CFG}} \mathcal{M}(\mathbf{M}^d_{\text{fut}}, d ; \mathbf{M}_{\text{pst}}, \mathbf{D}_{\text{pst}}, \mathbf{C}_{\text{fut}}, \mathbf{c}_{\text{syl}}) + (1 - \gamma_{\text{CFG}}) \mathcal{M}(\mathbf{M}^d_{\text{fut}}, d ; \varnothing, \varnothing, \mathbf{C}_{\text{fut}}, \mathbf{c}_{\text{syl}})
\end{split}
\end{equation}
where $\gamma_{\text{CFG}} \in [0, 1]$ defines the guidance strength, with lower values prioritizing style adherence over past-motion consistency. 

\subsubsection{Qualitative Results}
\label{sec:app_qual}
In this section, we show more qualitative results of our method.

\begin{figure}[!htpb]
    \centering
    \includegraphics[trim={15cm 7cm 0 18cm}, clip, width=\linewidth]{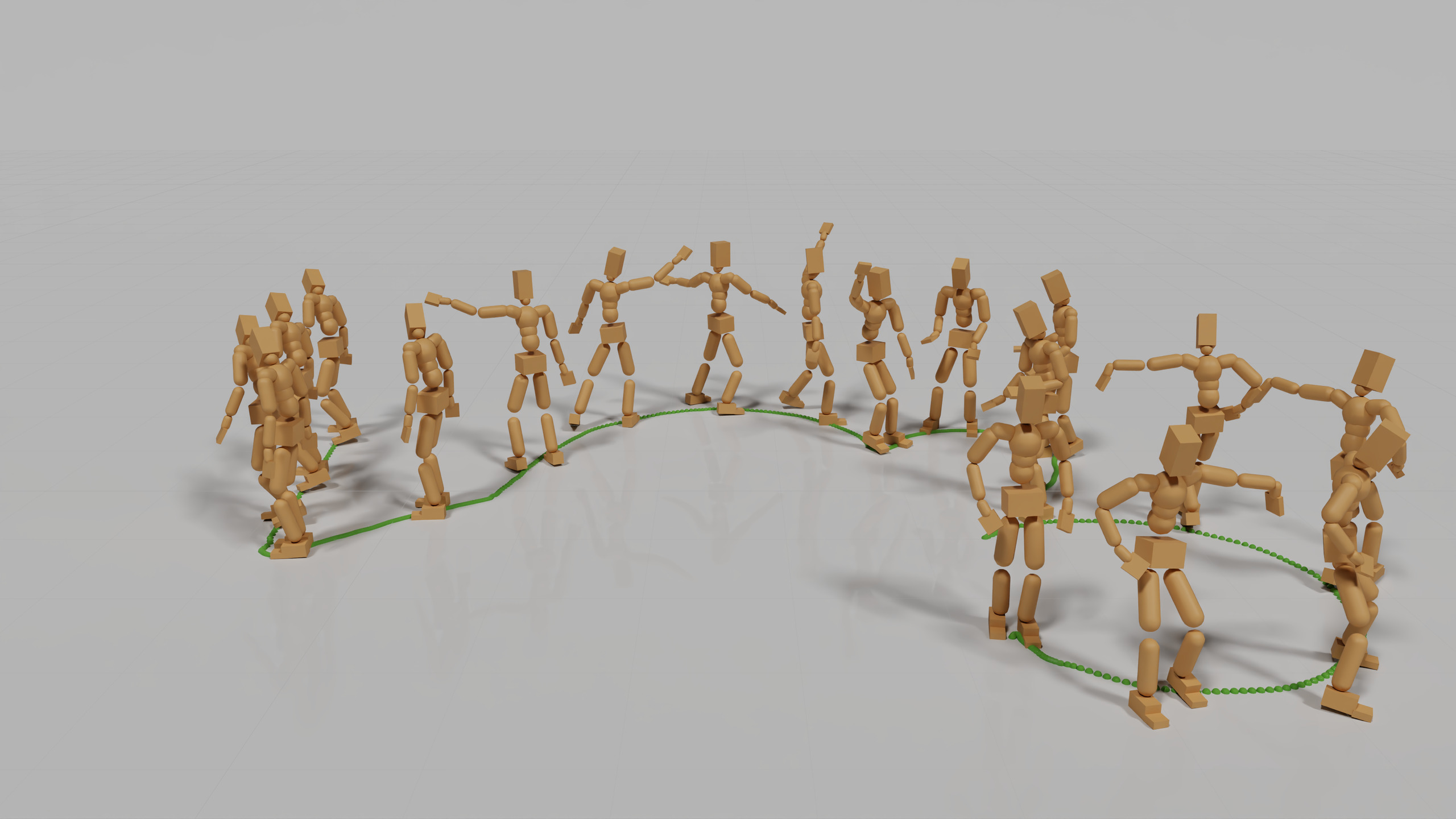}
    \caption{\textbf{Interactive Control.} Example of interactive control where a physics-based humanoid follows velocity commands in various styles. The green dots indicate the trajectory generated from the given velocity commands.}
    \label{fig:diffusion_tracked}
\end{figure}
\begin{figure}
    \centering
    \begin{minipage}{0.04\linewidth}
        \centering
        \rotatebox{90}{\textbf{CAMDM}}
    \end{minipage}%
    \begin{minipage}{0.96\linewidth}
        \includegraphics[trim={0cm 28cm 0cm 28cm}, clip, width=\linewidth]{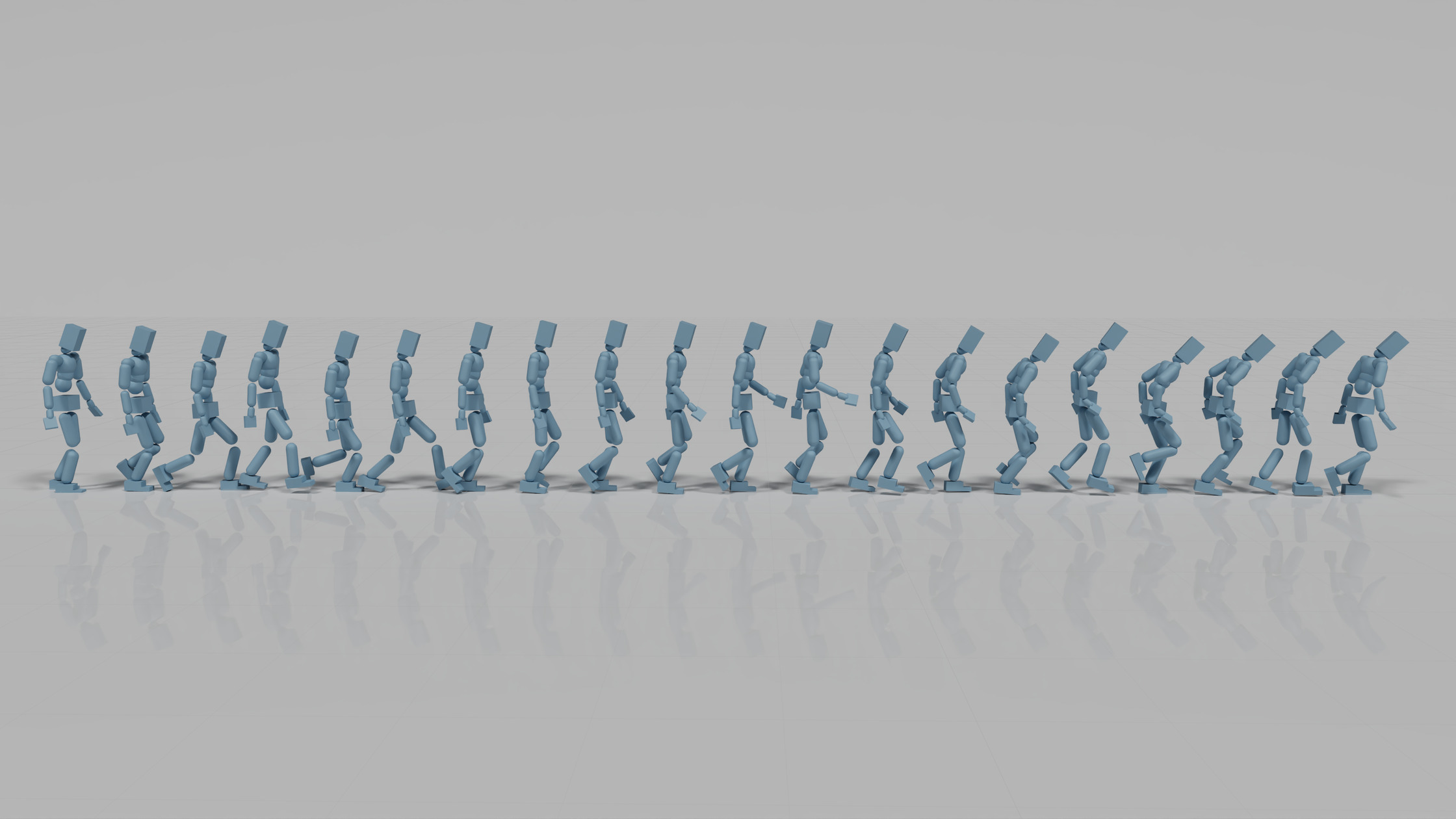}
    \end{minipage}

    \vspace{0.2cm} %

    \begin{minipage}{0.04\linewidth}
        \centering
        \rotatebox{90}{\textbf{Ours}}
    \end{minipage}%
    \begin{minipage}{0.96\linewidth}
        \includegraphics[trim={0cm 28cm 0cm 28cm}, clip, width=\linewidth]{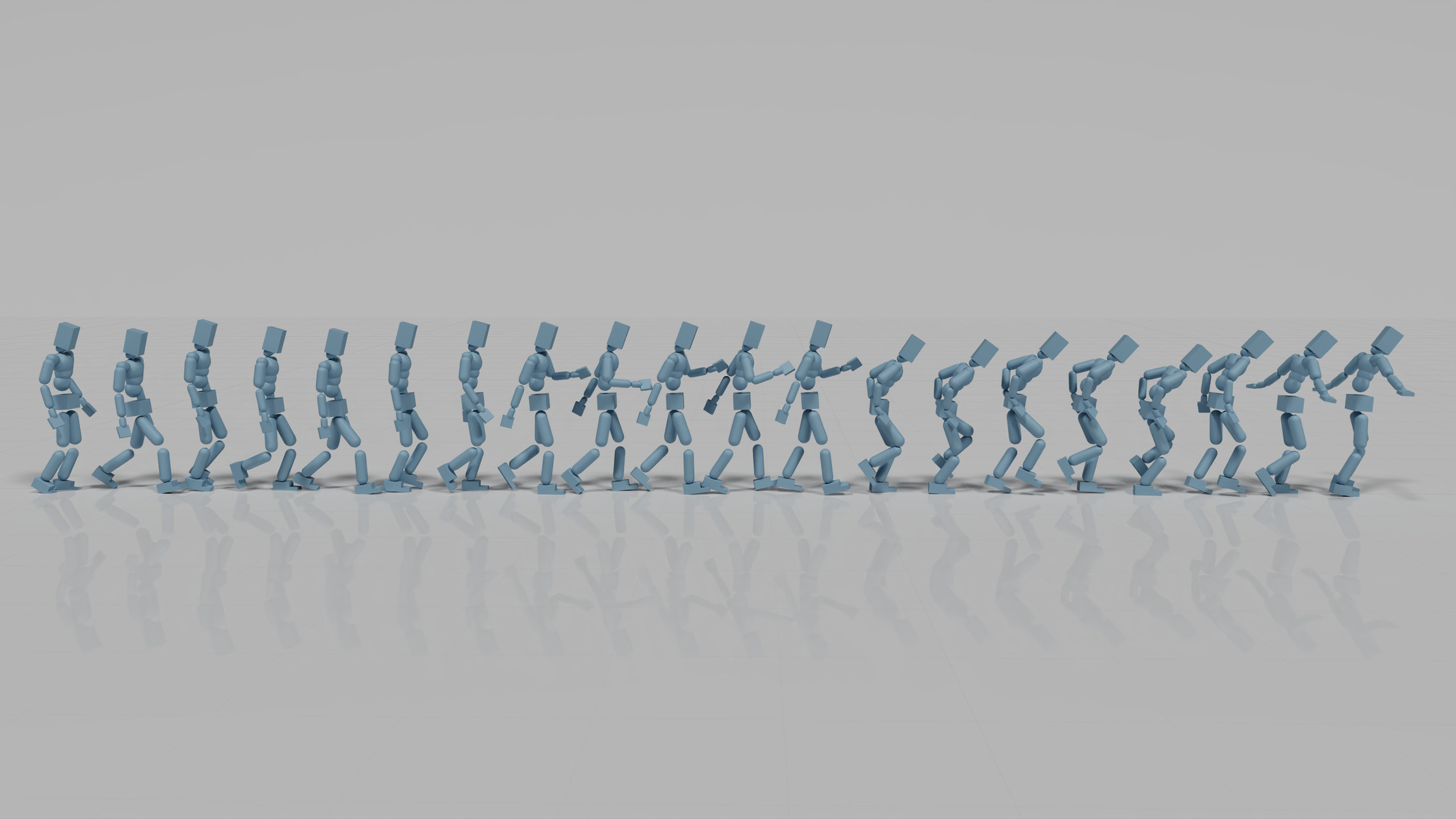}
    \end{minipage}
    
    \caption{\textbf{Baseline Comparison.} We show two motion sequences generated by the same commands with CAMDM \citep{chen_taming_2024} and our model.}
    \label{fig:ours_vs_camdm}
\end{figure}

\begin{figure}[!t]
    \centering
    \begin{tabular}{p{0.5\linewidth}@{\hspace{2pt}}p{0.5\linewidth}}
        \centering \textbf{Same window training} & \centering \textbf{Different window training (Ours)} \tabularnewline
        \includegraphics[trim={5cm 5cm 5cm 20cm}, clip, width=\linewidth]{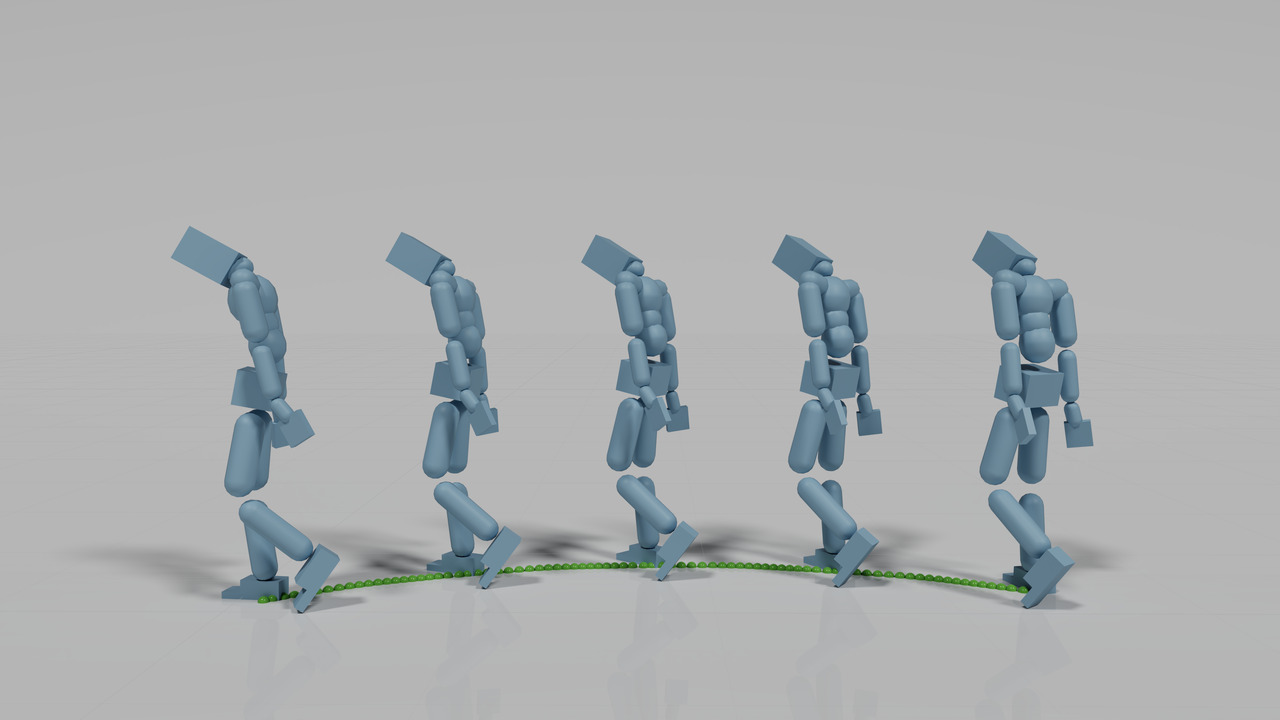} & 
        \includegraphics[trim={5cm 5cm 5cm 20cm}, clip, width=\linewidth]{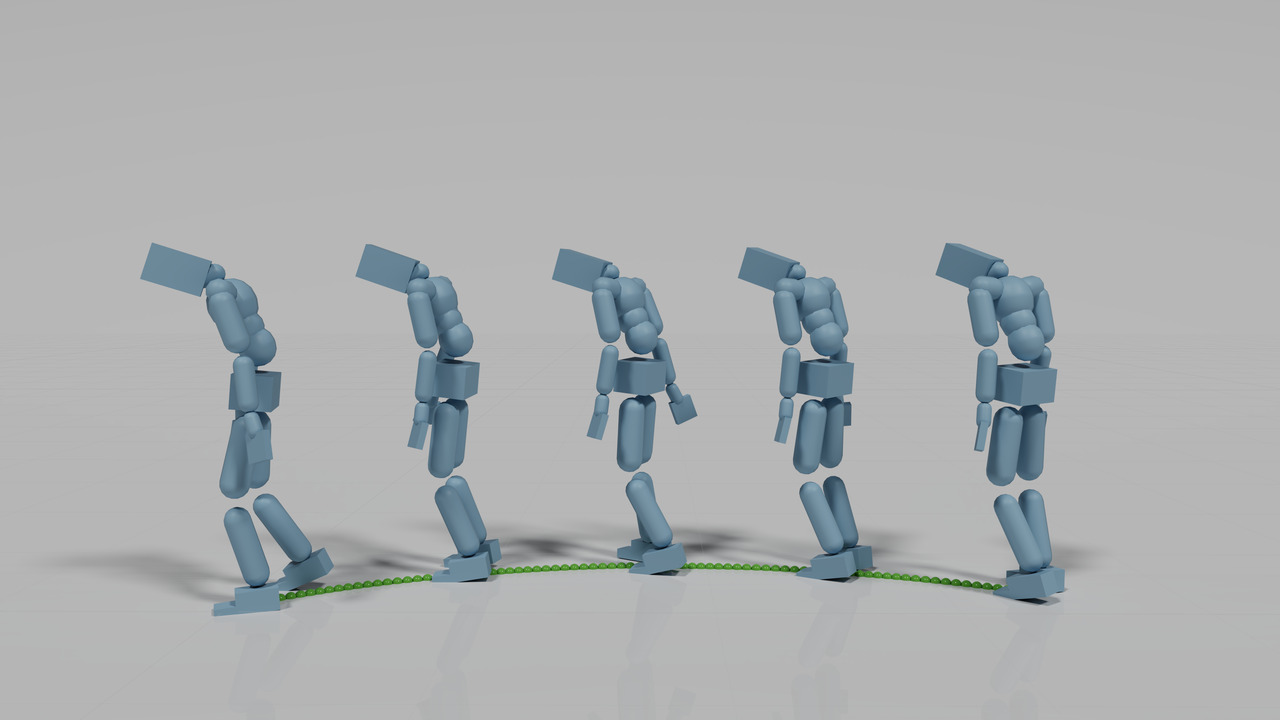}
    \end{tabular}
    
    \caption{\textbf{Sampling Scheme.} We show the first frame of 5 consecutively predicted motion windows of models trained with and without our proposed sampling scheme. Without our sampling approach (``same window training''), the model collapses and keeps repeating the reference motion. With our sampling approach (``different window training''), the model learns to extract the style characteristic from the reference motion.}
    \label{fig:sampling}
\end{figure}

\paragraph{Diffusion Model and Tracking} \Fig{diffusion_tracked} shows an example of a longer sequence of motion generated interactively. A user gives joystick and style commands as conditions for the diffusion model's generation which is then tracked by an RL policy.

\paragraph{Qualitative Comparison to CAMDM}
\Fig{ours_vs_camdm} shows a comparison of motions generated by \citep{chen_taming_2024} and our model for the same set of joystick commands and style conditions. For CAMDM we select the labels corresponding to the clips we choose as conditioning for our model. We can see that while the style is recognizable for both models, our method performs it more expressively, for example, by reaching the arms out widely in the aeroplane style (right).

\paragraph{Sampling Scheme}
We generate 5 consecutive predictions with and without our sampling scheme, given a fixed style reference as input. \Fig{sampling} shows how the model collapses without our sampling scheme, repeating the same motion at every prediction. With our sampling the model adapts and predicts a continuous motion that follows the general style characteristics.

\paragraph{Attention Bias}
\label{app:attention_bias}

\input{sections/figures/bias}
\begin{figure}[t!]
    \centering
    \includegraphics[width=\linewidth]{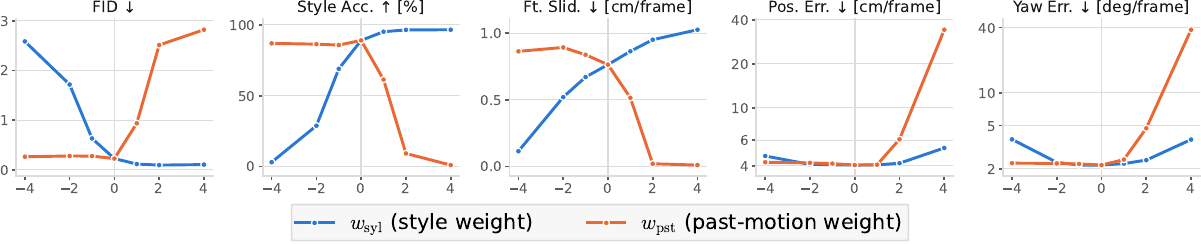}
    \caption{\textbf{Attention Bias Sensitivity.} We show how varying bias parameters affect performance by varying one weight at a time while keeping the other fixed at 0. Increasing either weight to large magnitudes worsens command adherence while increasing generation artifacts. Large positive $w_{\text{pst}}$ weights specifically result in frozen poses, which is why foot sliding is at zero.}
    \label{fig:bias_sweep}
\end{figure}

\Fig{attention_bias} shows qualitative comparisons of generating motions without any inference adjustments, with CFG and with our proposed attention biasing technique. Stylization is noticeably increased when using our biasing method. Additionally, \Fig{bias_sweep} shows how varying bias parameters affects performance. Increasing either weight's magnitude leads to worse command adherence, as the model is forced to attend less to the command conditions. Increasing $w_{\text{syl}}$ improves FID and style accuracy at the cost of increased foot sliding. Positive values of $w_{\text{pst}}$ reduce motion quality significantly, as they induce frozen poses, leading to foot sliding values of 0. Our defaults of $w_{\text{syl}} = 2.0$ and $w_{\text{pst}}=-1.0$ present a good trade-off between improved stylization quality and increased generation artifacts. We found these default values to work well across styles, embodiments and different models.

\begin{figure}[!tbp]
    \centering
    \includegraphics[width=\linewidth]{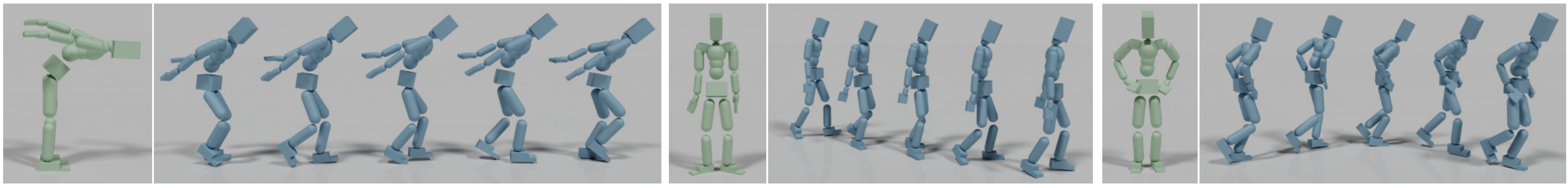}
    \caption{\textbf{Unseen Style.} Generation of motion with unseen styles from 100STYLE, shown as the green snapshots.}
    \label{fig:unseen}
\end{figure}

\paragraph{Unseen Styles}
To showcase generalization capabilities of our method, we select a random subset of the styles in 100STYLE as a test set (10 styles), which we do not include during training. Our model can still generate motion that matches these styles during inference, as shown in \Fig{unseen}.

\begin{figure}[!ht]
    \centering
    \includegraphics[width=\linewidth]{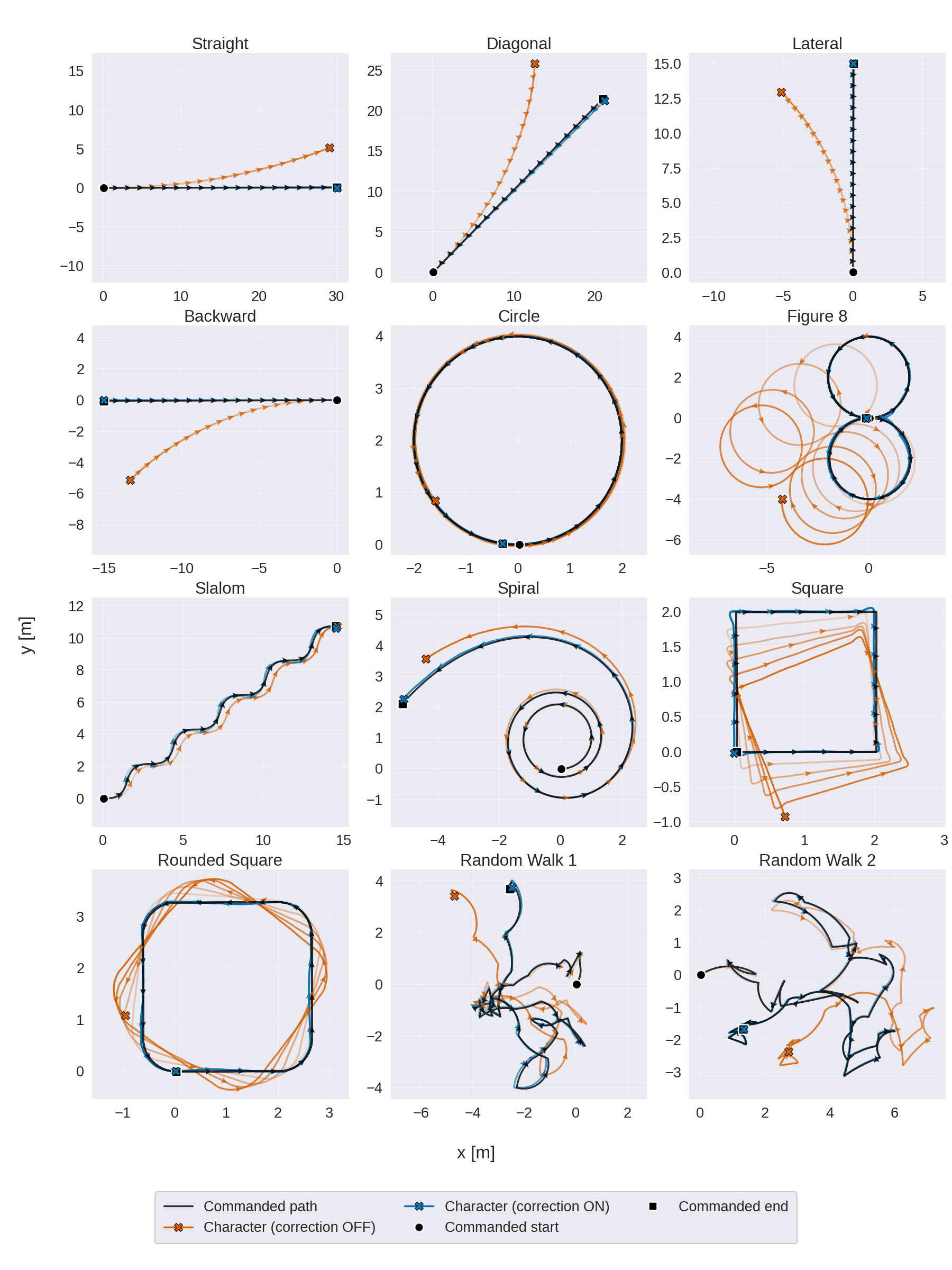}
    \caption{\textbf{Velocity Correction.} Qualitative analysis of applying velocity correction over a number of different trajectories.}
    \label{fig:velcorrect_qual}
\end{figure}

\paragraph{Velocity Correction} 
\Fig{velcorrect_qual} shows qualitative results of our assessment of the velocity correction mechanism. All trajectories correspond to roughly \SI{30}{\second} of motion. The periodic trajectories (circle, figure eights, square, rounded square) are repeated multiple times. In all cases, the character stays significantly closer to the commanded trajectory when the velocity correction scheme is applied.

%% file: sections/figures/smoothing.tex
\begin{figure}[tb]
    \centering
    \includegraphics[trim={0cm, 0cm, 0cm, 0cm}, clip, width=\linewidth]{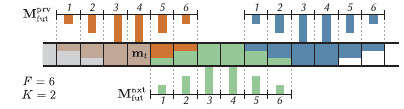}
    \caption{\textbf{Smoothing Scheme.} To achieve smooth transitions between consecutive diffusion calls, we predict the next $F$ states $\mathbf{M}_{\text{fut}}^{\text{nxt}}$ at the current timestep $t$, then blend the first $K$ states in $\mathbf{M}_{\text{fut}}^{\text{nxt}}$ with the last $K$ states of the previous prediction $\mathbf{M}_{\text{fut}}^{\text{prv}}$, then copy the next $F - 2 K$ states from $\mathbf{M}_{\text{fut}}^{\text{nxt}}$ before incrementing the timestep $t := t + (F-K)$ and repeating the process. More formally, we weight state $\mathbf{m}^{\text{nxt}}_{i}$ by $w = \nicefrac{i}{(K+1)}$ and state $\mathbf{m}^{\text{prv}}_{(F-K)+i}$ by $1 - w$ to form the blended states $\mathbf{m}_{t+i}$ for $i=1, \dots, K$. The visualization shows the $F=6, K=2$ case. }
    \label{fig:smoothing}
\end{figure}

%% file: sections/figures/bias.tex
\begin{figure}[t!]
    \centering
    \begin{tabular}{@{} c @{\hspace{2pt}} c @{\hspace{2pt}} c @{}}
        \textbf{Unmodified} & \textbf{CFG} & \textbf{Attention Bias} \\
        
        \includegraphics[trim = {0cm, 13cm, 21cm, 13cm}, clip, width=0.32\linewidth]{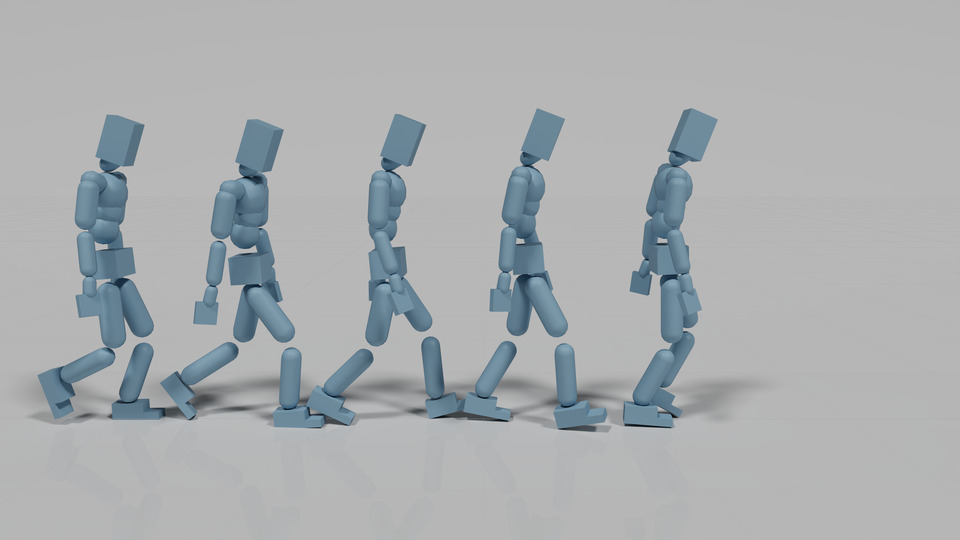} & 
        \includegraphics[trim = {0cm, 13cm, 21cm, 13cm}, clip, width=0.32\linewidth]{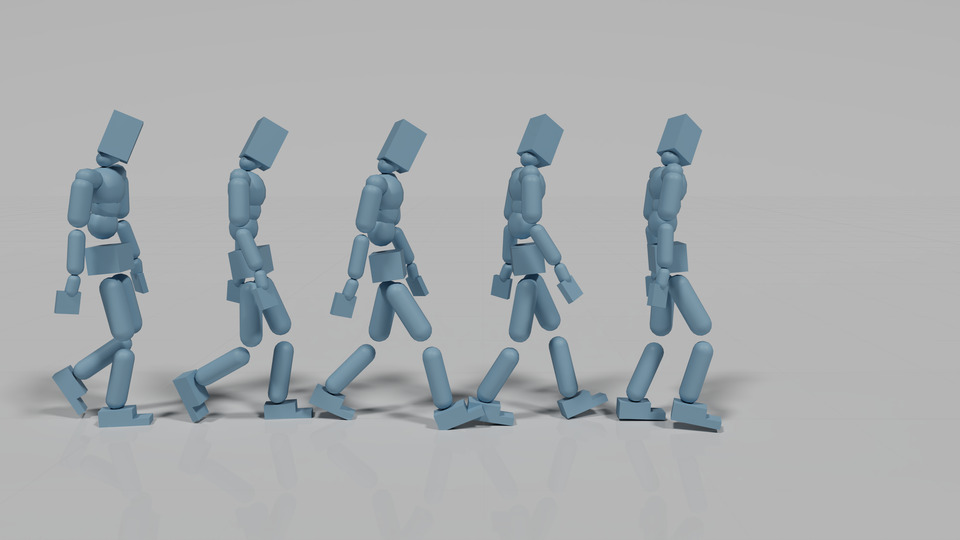} & 
        \includegraphics[trim = {0cm, 13cm, 21cm, 13cm}, clip, width=0.32\linewidth]{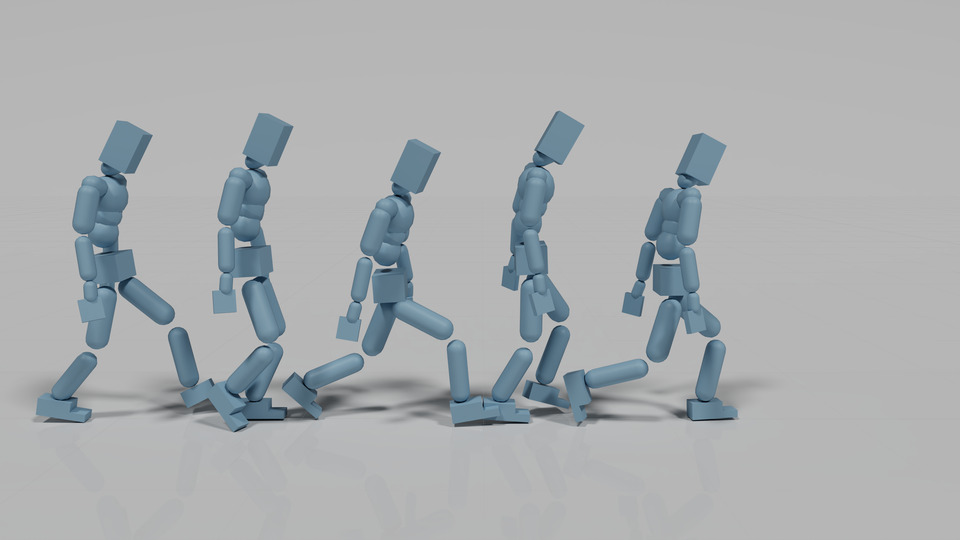} \\[-3pt]
        
        \includegraphics[trim = {0cm, 13cm, 21cm, 13cm}, clip, width=0.32\linewidth]{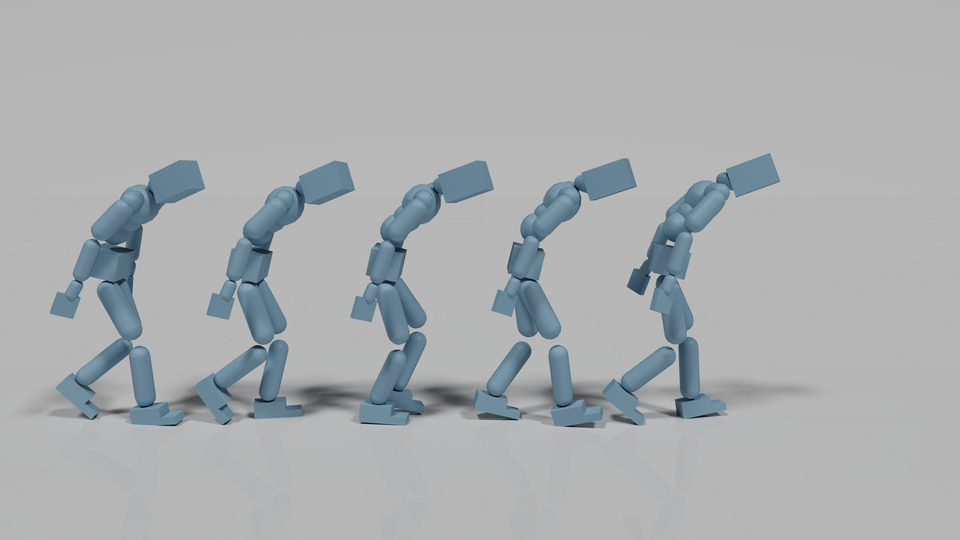} & 
        \includegraphics[trim = {0cm, 13cm, 21cm, 13cm}, clip, width=0.32\linewidth]{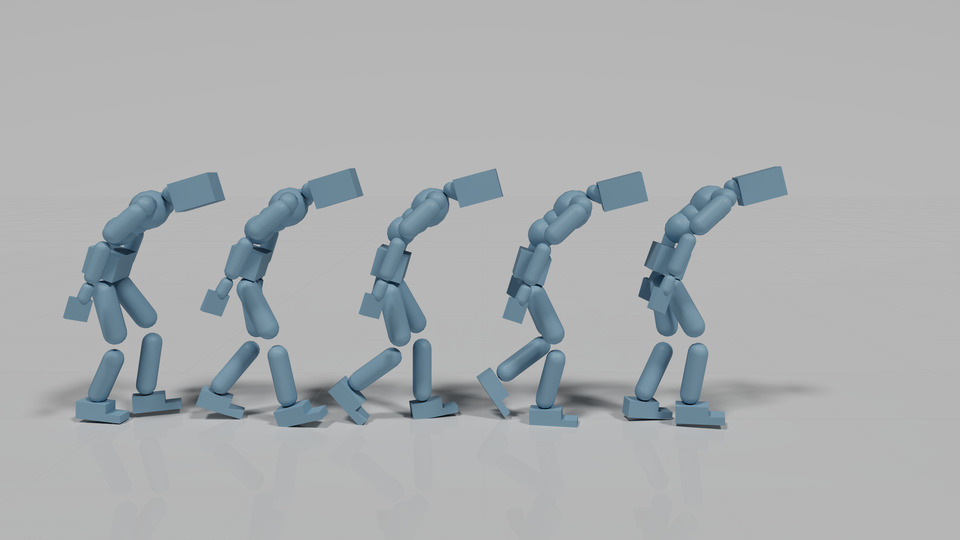} & 
        \includegraphics[trim = {0cm, 13cm, 21cm, 13cm}, clip, width=0.32\linewidth]{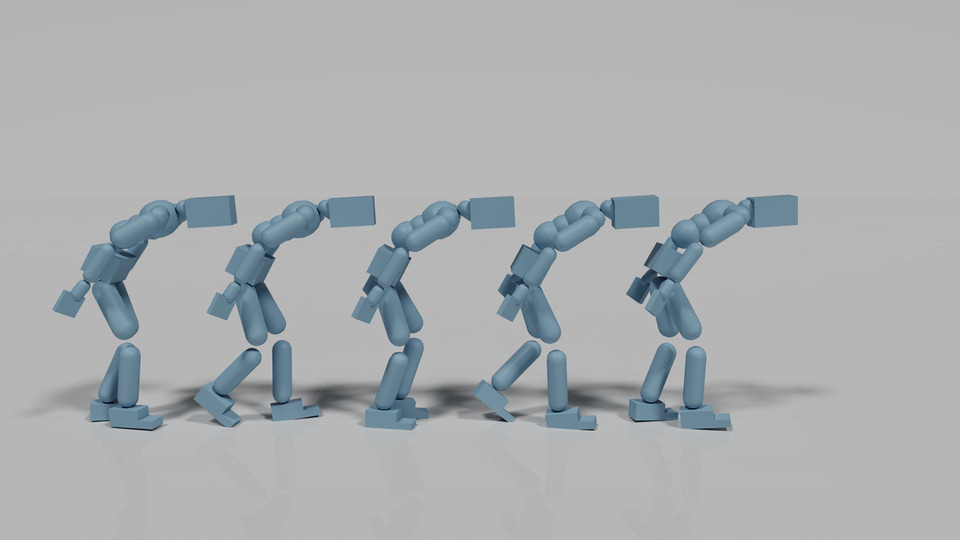} 
    \end{tabular}
    \caption{\textbf{Attention Bias Quality.} We compare unmodified inference with CFG and attention biasing for improved stylization.}
    \label{fig:attention_bias}
\end{figure}